\documentclass{article}

\usepackage{microtype}
\usepackage{graphicx}
\usepackage{subfigure}
\usepackage{booktabs} 

\usepackage{hyperref}

\usepackage[accepted]{icml2024}

\usepackage{url}
\usepackage{amsmath,bm}
\usepackage{mathrsfs}
\usepackage{multirow}
\usepackage{CJKutf8}
\usepackage{verbatim}
\usepackage{amssymb}
\usepackage{xspace}
\usepackage{enumitem}
\usepackage{makecell}
\usepackage{ragged2e}
\usepackage{multicol}

\usepackage{mathtools}
\usepackage{textcomp}
\usepackage{booktabs}       

\usepackage{amsthm}
\usepackage[capitalize,noabbrev]{cleveref}

\theoremstyle{plain}

\theoremstyle{definition}

\theoremstyle{remark}

\usepackage[textsize=tiny]{todonotes}

\icmltitlerunning{Language Generation with Strictly Proper Scoring Rules}

\DeclareUrlCommand\Code{\urlstyle{rm}}
\expandafter\def\expandafter\UrlBreaks\expandafter{\UrlBreaks  
\do\/\do\a\do\b\do\c\do\d\do\e\do\f\do\g\do\h\do\i\do\j\do\k
\do\l\do\m\do\n\do\o\do\p\do\q\do\r\do\s\do\t\do\u\do\v
\do\w\do\x\do\y\do\z
\do\A\do\B\do\C\do\D\do\E\do\F\do\G\do\H\do\I\do\J\do\K
\do\L\do\M\do\N\do\O\do\P\do\Q\do\R\do\S\do\T\do\U\do\V
\do\W\do\X\do\Y\do\Z}

\begin{document}

\twocolumn[
\icmltitle{Language Generation with Strictly Proper Scoring Rules}



\icmlsetsymbol{equal}{*}

\begin{icmlauthorlist}
\icmlauthor{Chenze Shao}{xxx}
\icmlauthor{Fandong Meng}{xxx}
\icmlauthor{Yijin Liu}{xxx}
\icmlauthor{Jie Zhou}{xxx}
\end{icmlauthorlist}

\icmlaffiliation{xxx}{Pattern Recognition Center, WeChat AI, Tencent Inc}

\icmlcorrespondingauthor{Chenze Shao}{chenzeshao@tencent.com}
\icmlcorrespondingauthor{Fandong Meng}{fandongmeng@tencent.com}
\icmlcorrespondingauthor{Yijin Liu}{yijinliu@tencent.com}
\icmlcorrespondingauthor{Jie Zhou}{withtomzhou@tencent.com}

\icmlkeywords{Machine Learning, ICML}

\vskip 0.3in
]



\printAffiliationsAndNotice{}  

\begin{abstract}
Language generation based on maximum likelihood estimation (MLE) has become the fundamental approach for text generation. Maximum likelihood estimation is typically performed by minimizing the log-likelihood loss, also known as the logarithmic score in statistical decision theory. The logarithmic score is strictly proper in the sense that it encourages honest forecasts, where the expected score is maximized only when the model reports true probabilities. Although many strictly proper scoring rules exist, the logarithmic score is the only local scoring rule among them that depends exclusively on the probability of the observed sample, making it capable of handling the exponentially large sample space of natural text. In this work, we propose a straightforward strategy for adapting scoring rules to language generation, allowing for language modeling with any non-local scoring rules. Leveraging this strategy, we train language generation models using two classic strictly proper scoring rules, the Brier score and the Spherical score, as alternatives to the logarithmic score. Experimental results indicate that simply substituting the loss function, without adjusting other hyperparameters, can yield substantial improvements in model's generation capabilities. Moreover, these improvements can scale up to large language models (LLMs) such as LLaMA-7B and LLaMA-13B. Source code: \url{https://github.com/shaochenze/ScoringRulesLM}.
\end{abstract}

\section{Introduction}

Language generation has played a pivotal role in the advancement of natural language processing, serving as the foundation for a wide range of applications \citep{NIPS2000_728f206c,Mikolov2010RecurrentNN,radford2018improving,brown2020language}. The primary goal of language generation is to learn the underlying probability distribution of a given text corpus. To achieve this, maximum likelihood estimation (MLE) is commonly employed to estimate the parameters of a probability distribution that best explains the text corpus \citep{myung2003tutorial}.

Maximum likelihood estimation is generally performed by minimizing the log-likelihood loss, also known as the logarithmic score, a prominent example of a strictly proper scoring rule \citep{good1952rational,gneiting2007strictly}. In statistical decision theory, scoring rules serve as quantitative measures to assess the quality of probabilistic predictions, by assigning a numerical score based on the predicted distribution and the observed sample. A scoring rule is considered strictly proper if it encourages models to report their true beliefs or probabilities. In other words, the expected score is maximized only when the model reports true probabilities, and any deviation from the truth will result in a lower expected score. Due to this property, strictly proper scoring rules are well-suited as loss functions for calibrating probabilistic models \citep{lakshminarayanan2017simple}. This is exemplified by the logarithmic score, which corresponds to the log-likelihood loss.

In addition to the logarithmic score, there are other strictly proper scoring rules that provide attractive loss functions for probabilistic prediction problems \citep{80304,hung1996estimating,Kline,hui2021evaluation}. However, only the logarithmic score has wide applications in language generation, primarily because it is the only strictly proper scoring rule that is also local: it depends exclusively on the predictive probability of the observed sample \cite{good1952rational,shuford1966admissible,bernardo1979expected}. Given the exponentially large sample space for natural text, calculating the score based on the entire probability distribution is infeasible, which hinders the application of non-local scoring rules in language modeling. Consequently, the logarithmic score, being both local and strictly proper, remains the only scoring rule capable of handling the exponentially large sample space of natural text. Nevertheless, the logarithmic score has faced criticism for its unbounded nature and sensitivity to small perturbations in the predicted distribution \citep{selten1998axiomatic}, suggesting that alternative strictly proper scoring rules might offer more suitable and robust options for training and evaluation in specific scenarios.

To investigate the impact and potential benefits of training language models with alternative strictly proper scoring rules, we propose a straightforward strategy for adapting non-local scoring rules to serve as loss functions for language generation. Specifically, we distribute the scoring rule at the token level to promote well-calibrated prediction of conditional probabilities at each time step, consequently leading to well-calibrated sequence-level probability predictions. We further introduce score smoothing to enable honest label smoothing for arbitrary scoring rules. Our approach allows language modeling with any non-local scoring rules while ensuring that the expected loss is minimized only when the model produces the desired probabilities. Leveraging this strategy, we train language generation models using two classic strictly proper scoring rules, the Brier score \citep{brier1950verification} and the Spherical score \citep{roby1965belief}, as alternatives to the logarithmic score. 

Experimental results indicate that simply substituting the loss function, without adjusting other hyperparameters, can yield substantial improvements in the model's generation capabilities. Moreover, these improvements can scale up to large language models (LLMs) such as LLaMA-7B and LLaMA-13B.

\section{Strictly Proper Scoring Rules}

In this section, we provide essential background on strictly proper scoring rules, including the definition and several popular examples.

\subsection{Scoring Rules}

Scoring rules assign a numerical score based on the predicted distribution and the observed sample. Let $\mathcal{X}=\{1,...,m\}$ represents the discrete sample space consisting of a finite number $m$ of different samples, and $\mathcal{P}_m=\{p=(p_1,...,p_m):p_1,...,p_m\geq0,\sum_{i=1}^{m}p_i=1\}$ be the set of probability measures on $\mathcal{X}$. A scoring rule $S(p,i)$ takes values in the extended real line $\overline{\mathbb{R}}=[-\infty,\infty]$, indicating the reward or utility of predicting $p$ when sample $i$ is observed:
\begin{equation}
 S(p,i):\mathcal{P}_m \times \mathcal{X} \mapsto \overline{\mathbb{R}}.
\end{equation}
Assuming samples conform to a data distribution $q$, we denote $S(p,q)$ as the expected score:
\begin{equation}
S(p,q)=\mathbb{E}_{i\sim q}[S(p,i)]=\sum_{i=1}^{m}q_i \cdot S(p,i).
\end{equation}
\subsection{Propriety}

A scoring rule is proper if the expected score is maximized when the model reports true probabilities:
\begin{equation}
S(p,q) \leq S(q,q), \ \ \ \  \forall p,q \in \mathcal{P}_m.
\end{equation}
It is strictly proper when the equality holds if and only if $p=q$. Propriety is an essential requirement for training and evaluating probabilistic models \citep{ScoringProbabilisticForecastsTheImportanceofBeingProper,lakshminarayanan2017simple}. In terms of training, strictly proper scoring rules can serve as training criteria to calibrate probabilistic models for well-calibrated prediction. In terms of evaluation, strictly proper scoring rules assess the quality of probabilistic predictions by measuring how they align with the true probabilities. 
\subsection{Locality}

A scoring rule is local if the probabilistic prediction is evaluated only at the observed sample, which means that there exists an equivalent function $S_{local}(p_i,i)$ that satisfies:
\begin{equation}
 S(p,i) = S_{local}(p_i,i), \ \ \ \  \forall p \in \mathcal{P}_m, i \in \mathcal{X}.
\end{equation}
A local scoring rule depends exclusively on the probability of the observed sample, rather than being rewarded for other features of the probabilistic distribution, such as its shape. It has been proven that every scoring rule being both proper and local is equivalent to the logarithmic score \citep{bernardo1979expected}. Formally, if $S$ is both proper and local, then for some constant $A$ and function $B$, we have:
\begin{equation}
S(p,i)=A\log p_i + B(i).
\end{equation}
\subsection{Examples}
\label{sec:examples}
We provide some examples of strictly proper scoring rules below.

\noindent{}\textbf{Logarithmic score.} The logarithmic score is a local scoring rule that measures the log probability of the observed sample. It is defined as:
\begin{equation}
S(p,i)=\log p_i.
\end{equation}
This scoring rule is closely related to maximum likelihood estimation and is widely used in language modeling. Despite its widespread use, the logarithmic score has been criticized for being unbounded and sensitive to small perturbations in the predicted distribution \citep{selten1998axiomatic}.

\noindent{}\textbf{Brier score.} The Brier score \citep{brier1950verification} is a quadratic scoring rule that measures the mean squared difference between the predicted distribution and the true outcome. It is defined as:
\begin{equation}
S(p,i)=1-\sum_{j=1}^{m}(\delta_{ij}-p_j)^2=2p_i-\sum_{j=1}^{m}p_j^2,
\end{equation}
where $\delta_{ij}=1$ if $i=j$ and $\delta_{ij}=0$ otherwise. The expected Brier score is $S(p,q)=\sum_{i=1}^{m}q_i^2-(p_i-q_i)^2$, which is maximized when $p=q$. A more general form is the $\alpha$-power score \citep{selten1998axiomatic}:
\begin{equation}
S(p,i)=\alpha p_i^{\alpha-1}-(\alpha-1)\sum_{j=1}^{m}p_j^{\alpha},\ \ \alpha > 1.
\end{equation}
The $\alpha$-power score defines a family of strictly proper scoring rules, with the Brier score being a special case for $\alpha=2$.

\noindent{}\textbf{Spherical score.} The spherical score \citep{roby1965belief} measures the cosine similarity between the predicted probability vector and the true probability vector. It is defined as:
\begin{equation}
S(p,i)=\frac{p_i}{|p|}.
\end{equation}
The expected spherical score, $S(p,q)={\langle p, q \rangle}/{|p|}$, is proportional to the cosine similarity and is therefore maximized when $p=q$. A more general form is the pseudo-spherical score:
\begin{equation}
S(p,i)=\frac{p_i^{\alpha - 1}}{(\sum_{j=1}^m p_j^{\alpha})^{\frac{\alpha-1}{\alpha}}}, \ \ \alpha > 1.
\end{equation}
It reduces to the spherical score when $\alpha = 2$. Note that both the $\alpha$-power score and the pseudo-spherical score depend on the current prediction probability $p_i$ as well as the global characteristics of the distribution, i.e., the $\alpha$-norm of $p$. Therefore, they are strictly proper but non-local.

In addition to the classic scores introduced above, strictly proper scoring rules can also be constructed from any bounded strictly convex function on $\mathcal{P}_m$. Please refer to \citet{gneiting2007strictly} for a literature review.

\section{Language Generation with Strictly Proper Scoring Rules}

In this section, we present our strategy for adapting non-local scoring rules to serve as loss functions for language generation. Section \ref{sec:31} introduces the framework of utilizing scoring rules as loss functions. Section \ref{sec:32} describes our approach for distributing the scoring rule at the token level, which overcomes the locality constraint. Section \ref{sec:33} further adapts scoring rules to support regularization with label smoothing.

For simplicity of notation, we focus on unconditional sequence models in this section, where samples $x\in \mathcal{X}$ consist of discrete tokens $x=\{x_1,x_2,...,x_T\}$. The data distribution is represented by $q(x)$, the model predicts the distribution $p_{\theta}(x)$, and the scoring rule is denoted as $S(p_{\theta},x)$. The subsequent discussion can be directly extended to conditional sequence generation scenarios, such as translation and summarization tasks.
\subsection{Scoring Rules as Losses}
\label{sec:31}
Scoring rules assign a numerical score based on the predicted distribution $p_{\theta}$ and the observed sample $x$, which can be interpreted as the reward or utility of predicting $p_{\theta}$ when sample $x$ is observed. It is natural to maximize the scoring rule $S$ by minimizing the associated loss function $\mathcal{L}_S$:
\begin{equation}
\mathcal{L}_S(\theta)=-S(p_{\theta},q)=-\mathbb{E}_{x\sim q} S(p_{\theta},x).
\end{equation}
As long as $S$ is strictly proper, the associated loss $\mathcal{L}_S$ will have a unique minimizer $p_{\theta}=q$, encouraging the model to report the true distribution $q$.

In sequence prediction problems, given the maximum length $T_{max}$ and vocabulary size $V$, the sample space has an exponentially large size of $V^{T_{max}}$. This makes it intractable to calculate scoring rules that depend on global characteristics of the distribution, such as the Brier score and the spherical score. The logarithmic score, being both local and strictly proper, remains the only scoring rule capable of handling sequence prediction problems. The corresponding loss function is:
\begin{equation}
\label{eq:log}
\mathcal{L}_{log}(\theta)=-\mathbb{E}_{x\sim q} \log p_{\theta}(x).
\end{equation}
This loss function can also be derived from maximum likelihood estimation and is commonly referred to as the log-likelihood loss or cross-entropy loss.

\subsection{Token-Level Scoring Rules}
\label{sec:32}
In general, sequence models do not directly compute the probability of entire sequences. Instead, they decompose the sequence probability into a product of token probabilities in an autoregressive manner:
\begin{equation}
p_{\theta}(x)=\prod_{t=1}^{T}p_{\theta}(x_t|x_{<t}).
\end{equation}
This autoregressive decomposition transforms the sequence prediction task into a series of conditional token prediction tasks, where the sample space is reduced to $V$ for each task. As long as the model predicts the accurate conditional token probability $q(x_t|x_{<t})$, it can correctly recover the sequence probability $q(x)$. Therefore, we can distribute the scoring rule at the token-level to promote well-calibrated prediction for each token prediction task. In this way, we define the following loss based on token-level scoring rules:
\begin{equation}
\begin{aligned}
\mathcal{L}&_{S}(\theta)=-\mathbb{E}_{x\sim q}[\sum_{t=1}^{T} S(p_{\theta}(\cdot|x_{<t}),x_t)]\\
&=-\sum_{t=1}^{T} \mathbb{E}_{x_{<t}\sim q}[\sum_{x_t}q(x_t|x_{<t})S(p_{\theta}(\cdot|x_{<t}), x_t)]\\
&=-\sum_{t=1}^{T} \mathbb{E}_{x_{<t}\sim q}[S(p_{\theta}(\cdot|x_{<t}), q(\cdot|x_{<t}))].
\end{aligned}
\end{equation}
In the above equation, $p_{\theta}(\cdot|x_{<t})$ and $q_{\theta}(\cdot|x_{<t})$ are probability vectors of size $|V|$, representing the conditional probability distributions of the next word given the history $x_{<t}$. The equation shows that the loss is minimized only when each token-level scoring rule $S(p_{\theta}(\cdot|x_{<t}), q(\cdot|x_{<t}))$ is maximized. For strictly proper $S$, maximizing the score means matching every $p_{\theta}(\cdot|x_{<t})$ with $q(\cdot|x_{<t})$, consequently leading to well-calibrated probability predictions $p_{\theta}=q$:
\begin{equation}
p_{\theta}(x)=\prod_{t=1}^{T}p_{\theta}(x_t|x_{<t})=\prod_{t=1}^{T}q(x_t|x_{<t})=q(x).
\end{equation}

Token-level score optimization allows for language modeling with any non-local strictly proper scoring rules, such as the Brier score \citep{brier1950verification} and the spherical score \citep{roby1965belief}. For the Brier score, the token-level loss is given by:
\begin{equation}
\mathcal{L}_{Brier}(\theta)=-\mathbb{E}_{x\sim q} \sum_{t=1}^{T} 2p_{\theta}(x_t|x_{<t})-|p_{\theta}(\cdot|x_{<t})|^2.
\end{equation}
The token-level loss for the spherical score is:
\begin{equation}
\mathcal{L}_{Spherical}(\theta)=-\mathbb{E}_{x\sim q} \sum_{t=1}^{T} \frac{p_{\theta}(x_t|x_{<t})}{|p_{\theta}(\cdot|x_{<t})|}.
\end{equation}
For the logarithmic score, its token-level loss formulation is equivalent to the sequence-level one defined in Equation \ref{eq:log}.
\subsection{Score Smoothing}
\label{sec:33}
In practical applications, it is not always expected for a model to perfectly fit the data distribution, as the label smoothing technique \cite{szegedy2016rethinking} might be employed for regularization purposes. Label smoothing is typically used in classification and sequence prediction tasks, where it modifies the cross-entropy loss by replacing the one-hot label vector with a soft label to avoid overconfident predictions. 

Suppose we have a label set $\mathcal{X}=\{1,...,m\}$ and a label distribution $q$. Label smoothing with a smoothing factor $\epsilon$ encourages the model to produce a smooth distribution $q^\epsilon$:
\begin{equation}
q_i^\epsilon=(1-\epsilon)q_i + \frac{\epsilon}{m}.
\end{equation}

The current label smoothing technique is limited to loss functions based on the logarithmic score. Here we introduce a general smoothing technique called score smoothing, which supports label smoothing for arbitrary scoring rules. Given a smoothing factor $\epsilon$, the smoothed score $S^\epsilon$ is defined as:
\begin{equation}
S^\epsilon(p,i)=(1-\epsilon) \cdot S(p,i) + \frac{\epsilon}{m} \cdot \sum_{j=1}^{m}S(p,j).
\end{equation}
Ideally, score smoothing should be consistent with the goal of label smoothing, motivating the model to generate the desired smooth distribution $q^\epsilon$. In this context, we define a smoothed score as proper if the expected score satisfies $S^\epsilon(p,q)\leq S^\epsilon(q^\epsilon,q)$, and it is strictly proper when the equality only holds at $p=q^\epsilon$. The following equation shows that $S^\epsilon$ is strictly proper as long as $S$ is strictly proper:
\begin{equation}
\begin{aligned}
S^\epsilon(p,q)&=(1-\epsilon)  \sum_{i=1}^{m}q_i S(p,i) + \frac{\epsilon}{m}  \sum_{i=1}^{m}S(p,i)\\
&=\sum_{i=1}^{m} ((1-\epsilon)q_i + \frac{\epsilon}{m}) \cdot S(p,i)\\
&=S(p,q^{\epsilon}).
\end{aligned}
\end{equation}
By definition, the expected smoothed score $S^\epsilon(p,q)=S(p,q^{\epsilon})$ is maximized only when the model produces $q^{\epsilon}$, proving that $S^\epsilon$ is strictly proper as well.

However, when applying score smoothing in practice, we observe that the smoothing term might be ignored in some scoring rules. This is primarily attributed to the corresponding loss being relatively flat around the optimal point $q^{\epsilon}$. Since a nearly equivalent minimal loss can be achieved without taking the smoothing term into account, the model lacks enough incentive to perform label smoothing. 

Consider an example with the number of labels $m=100$, a one-hot true probability $q=(1,0,0,...,0)$, and a smoothing factor $\epsilon=0.1$. Table \ref{tab:example} gives the expected score $S^\epsilon(p,q)=S(p,q^{\epsilon})$ when the model produces $p=q,q^{\epsilon}$ respectively. The logarithmic score imposes $-\infty$ score penalty for disregarding score smoothing with $p=q$. In contrast, the Brier score and the spherical score are bounded, which can only impose a relatively mild penalty when score smoothing is ignored. In particular, the spherical score exhibits nearly identical expected scores in both cases, causing the smoothing term to be almost disregarded.

\begin{table}[t]
\caption{Expected scores when the model conducts or ignores score smoothing.}
\label{tab:example}
\vskip 0.1in
\begin{center}
\begin{small}
\begin{tabular}{lccc}
\toprule
{\bf Score} &{\bf Logarithmic }&{\bf Brier} & {\bf Spherical }\\
\midrule
$S(p=q,q^{\epsilon})$ & $-\infty$  &0.8020 & 0.9010 \\
\midrule
$S(p=q^{\epsilon},q^{\epsilon})$ & -0.7778 & 0.8119 & 0.9011 \\
\bottomrule
\end{tabular}
\end{small}
\end{center}
\vspace{-1em}
\end{table}

To address this limitation, we introduce a masked logarithmic score to enhance the smoothing effect. In the target distribution $q^{\epsilon}$, all labels have a probability of at least $\frac{\epsilon}{m}$. Therefore, labels with probabilities below this threshold can be considered under-smooth. We apply the logarithmic score to further augment the smoothing term for these under-smooth labels:
\begin{equation}
S^\epsilon_{\log}(p,i) = S^\epsilon(p,i) + \frac{\epsilon}{m} \sum_{j=1}^{m} 1\{p_j < \frac{\epsilon}{m}\} \log p_j,
\end{equation}
where $1\{\cdot\}$ is the indicator function that takes the value $1$ if the inside condition holds. Since the logarithmic score is only applied to under-smooth labels, it does not affect the propriety of the score. Formally, for strictly proper $S$, we have:
\begin{equation}
S^\epsilon_{\log}(p,q) \leq S^\epsilon(p,q) \leq S^\epsilon(q^{\epsilon},q) = S^\epsilon_{\log}(q^{\epsilon},q).
\end{equation}
Therefore, the expected score is maximized only when $p=q^\epsilon$, implying that $S^\epsilon_{\log}$ is strictly proper. Enhanced by the masked logarithmic score, it ensures a stronger incentive for the model to produce the desired smooth distribution $q^{\epsilon}$.

\begin{figure*}[t]
\centering
\begin{minipage}{.33\textwidth}
  \centering
  \includegraphics[width=1.\linewidth]{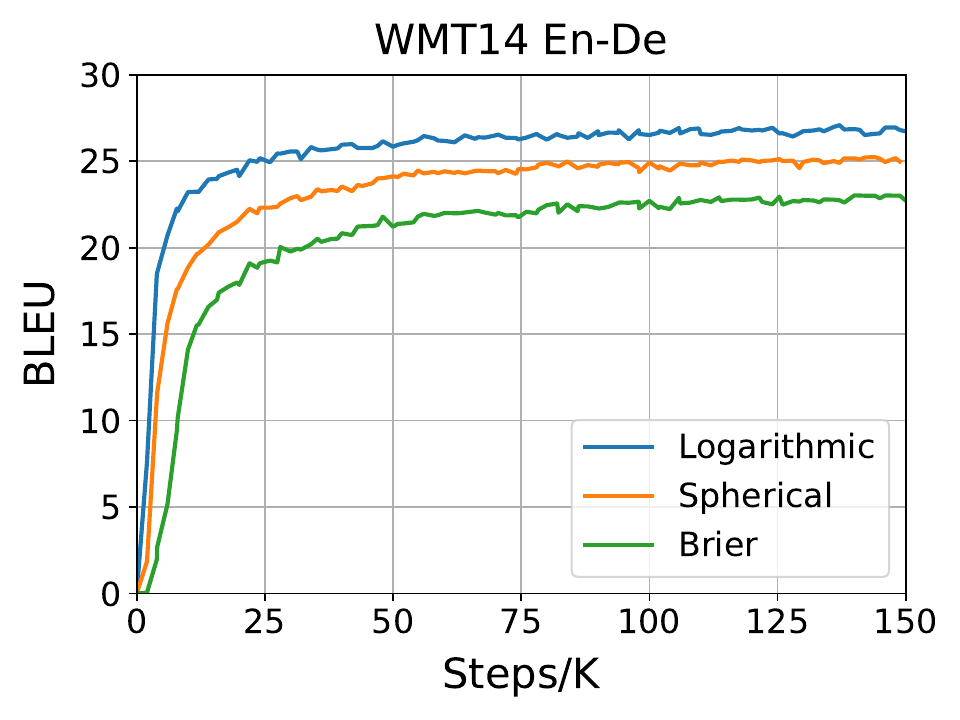}
\end{minipage}
\begin{minipage}{.33\textwidth}
  \centering
  \includegraphics[width=1.\linewidth]{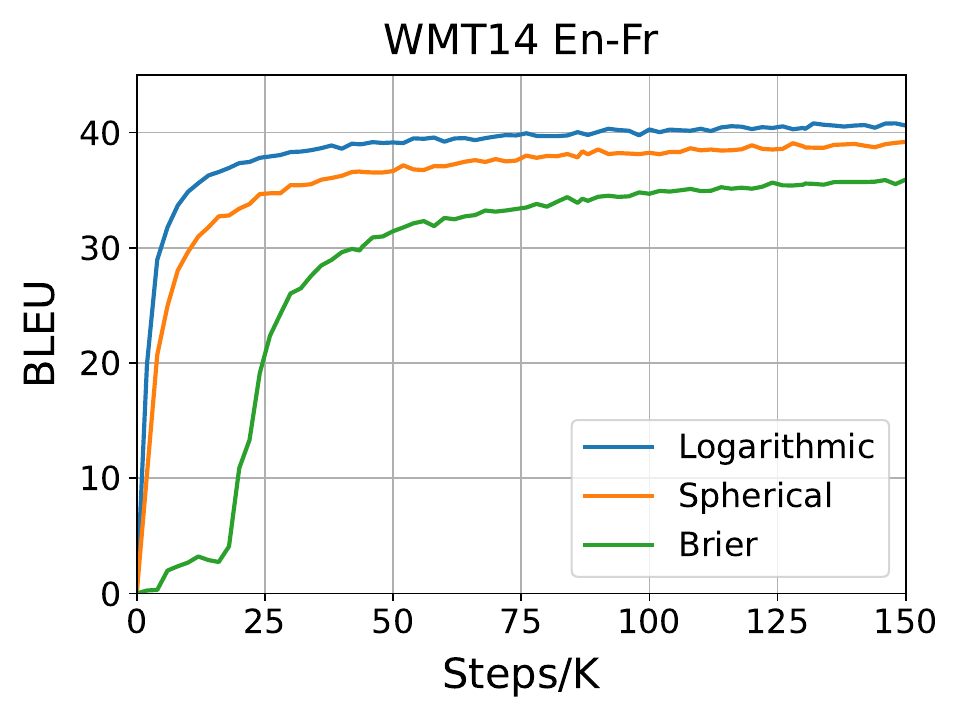}
\end{minipage}
\begin{minipage}{.33\textwidth}
  \centering
  \includegraphics[width=1.\linewidth]{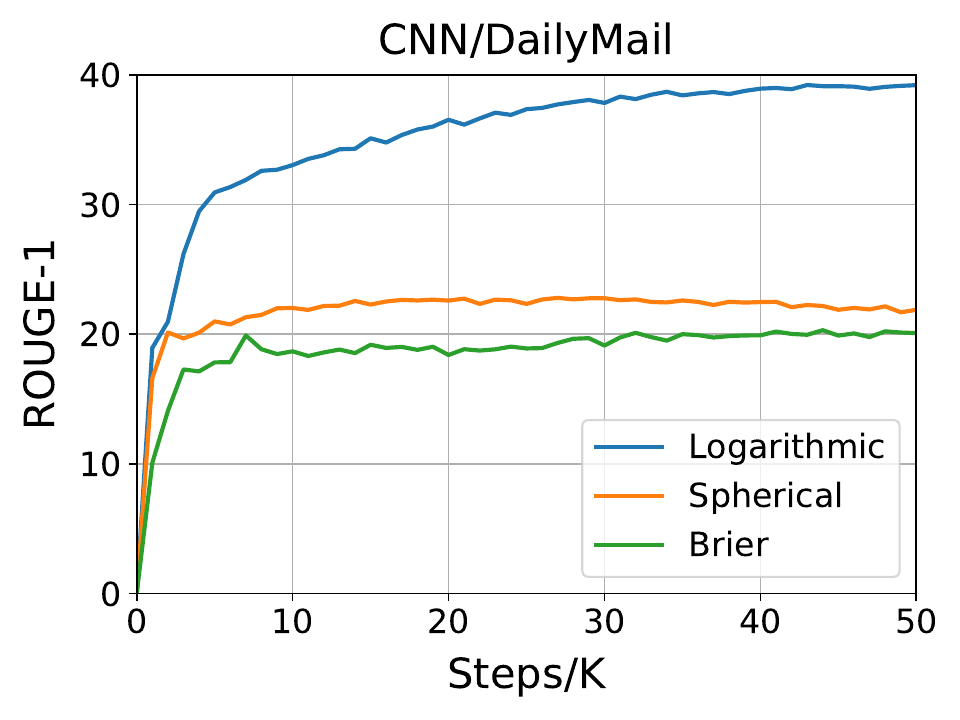}
\end{minipage}
  \caption{Performance curves of different strictly proper scoring rules on translation and summarization tasks.}
  \label{fig:scratch}
  \vspace{-0.5em}
\end{figure*}
\section{Experiments}

\subsection{Setup}

\begin{table}[t]
\caption{Implementation details on different datasets.}
\label{table:Table_params}
\vskip 0.1in
\begin{center}
\begin{small}
\begin{tabular}{lcccc}
\toprule
Dataset&  En-De &  En-Fr & TED & CNN \\
\midrule
 batch size & 32k & 32k & 32k & 64k  \\
 learning rate &7e-4 & 5e-4 &7e-4 &2e-4   \\
 dropout & 0.1  &  0.1 & 0.3 & 0.1    \\
 attention dropout  & 0  &  0 & 0 & 0.1    \\
 warmup steps   &4k & 4k &4k &2k  \\
 training steps &200k & 300k & 18k &100k  \\
 fine-tuning steps &50k & 50k & 4k &20k  \\
 weight decay &0 & 0 & 0.0 & 0.01  \\
 beam size & 5 & 5 & 5 & 4  \\
 length penalty &0 & 0.6& 1&2 \\
\bottomrule 
\end{tabular}
\end{small}
\end{center}
\vspace{-1em}
\end{table}

\textbf{Datasets.} We primarily evaluate our method on machine translation and abstractive summarization. For machine translation, we conduct experiments on widely used translation benchmarks under difference scales: WMT14 English-French (En-Fr, 35.8M pairs), WMT14 English-German (En-De, 4.5M pairs), TED bilingual dataset (10 directions, each with $\sim$200K pairs). For WMT datasets, we use \textit{newstest2013} for validation and \textit{newstest2014} for test, and apply BPE \citep{DBLP:conf/acl/SennrichHB16a} with 32K merge operations to learn a joint vocabulary on the tokenized data. For TED bilingual dataset, we use the pre-processed data used in \citet{xu-etal-2021-vocabulary}. The translation quality is measured by BLEU \citep{papineni-etal-2002-bleu}. For abstractive summarization, We conduct experiments on the summarization benchmark CNN/DailyMail \citep[311K pairs,][]{NIPS2015_afdec700}. We use the non-anonymized variant following \citet{see-etal-2017-get,DBLP:conf/acl/LiuYGQZJCFSGWCJ21}. The summarization quality is measured with ROUGE-1, ROUGE-2, and ROUGE-L \citep{lin-2004-rouge}. We adopt the settings of Transformer-base \citep{DBLP:conf/nips/VaswaniSPUJGKP17} for most datasets, except that we use Transformer-big for WMT14 En-Fr. Implementation details are provided in Table \ref{table:Table_params}. 

\noindent{}\textbf{Large Language Models.} We further investigate the performance of scoring rules at a larger model scale. Due to the large computational cost of pre-training, we utilize two open-source large language models \citep[LLaMA-7B and LLaMA-13B,][]{touvron2023llama} as our foundation models, and only employ strictly proper scoring rules for instruction tuning. We conduct instruction tuning using the Alpaca dataset by GPT4 \cite{selfinstruct,alpaca}, which comprises 52K instruction-following demonstrations. We keep the standard settings for instruction tuning on Alpaca, except that the log-likelihood loss is replaced with losses associated with other scoring rules.

Similarly, the generative capability of LLMs is evaluated on machine translation and abstractive summarization. Besides, we also employ MT-bench, a multi-turn question set, to evaluate the open-ended question answering capabilities of LLMs. For machine translation, we follow previous works \cite{jiao2023parrot,bayling,zeng2023tim,liu2023instruction} to evaluate the translation capability on four WMT22 translation tasks (Chinese-to-English, English-to-Chinese, German-to-English, and English-to-German). For text summarization, we follow \citet{liu2023instruction} to conduct the evaluation on CNN/DailyMail Dataset. We employ beam search with a beam size of 4 for machine translation and 2 for summarization. The prompt for machine translation is "Translate the following sentences from [SRC] to [TGT]." The prompt for summarization is "Write a brief and focused summary of the passage that follows.".

\subsection{Training from Scratch}

In our initial investigation, we evaluate the performance of various strictly proper scoring rules when training language generation models from scratch. We employ three typical scoring rules - the logarithmic score, the Brier score, and the spherical score - to train language generation models. Figure \ref{fig:scratch} displays their performance curves on three datasets: WMT14 En-De, WMT14 En-Fr, and CNN/DailyMail. 

The results indicate that, although all of these scoring rules are strictly proper, they still exhibit noticeable differences when training language generation models from scratch. Among the three datasets, the logarithmic score consistently converges the fastest and achieves the best performance. The spherical score follows, and the Brier score exhibits the slowest convergence and the lowest performance.

We hypothesize that such differences may be attributed to two primary factors. On one hand, despite sharing the same optimum of $p=q$, different strictly proper scoring rules possess distinct learning dynamics. For a specific neural architecture, the optimization trajectory and achievable optimum for each score vary, depending on the characteristics of the score's gradient. For instance, compared to the other two scores, the logarithmic score exhibits a larger gradient during the initial stages of training, which may facilitate model warmup and enable faster convergence. On the other hand, the hyperparameter settings we employed were obtained from previous works that used the logarithmic score for training \citep{DBLP:conf/nips/VaswaniSPUJGKP17}. These settings may not be as well-suited for other scoring rules, resulting in their relatively inferior performance.

\begin{table}[t]
\caption{BLEU scores on WMT14 En-De and WMT14 En-Fr test sets. `+ Brier' and `+ Spherical' represent fine-tuning with the Brier score or the Spherical score. The compared methods are based on our implementation. Statistical significance is indicated by $^*(p < 0.01)$ vs. the baseline. }
\label{table:Results_mt}
\vskip 0.1in
\begin{center}
\begin{small}
\begin{tabular}{lccc}
\toprule
{\bf Model}  & \bf EN-DE & \bf EN-FR \\
\midrule
Transformer  & 27.61 & 41.92 \\
\midrule
MixCE \citep{zhang-etal-2023-mixce}& 27.75 & 42.03 \\
TaiLr \citep{ji2023tailoring} & 27.95& 42.12 \\ 
Convex \citep{shao2023beyond} & 27.80 & 42.05 \\
\midrule
Transformer + Brier & 28.01$^{*}$ & \bf 42.50$^{*}$ \\
Transformer + Spherical & \bf 28.07$^{*}$ & 42.09  \\
\bottomrule 
\end{tabular}
\end{small}
\end{center}
\vspace{-1em}
\end{table}
\begin{figure}[t]
  \begin{center}
    \includegraphics[width=1.\columnwidth]{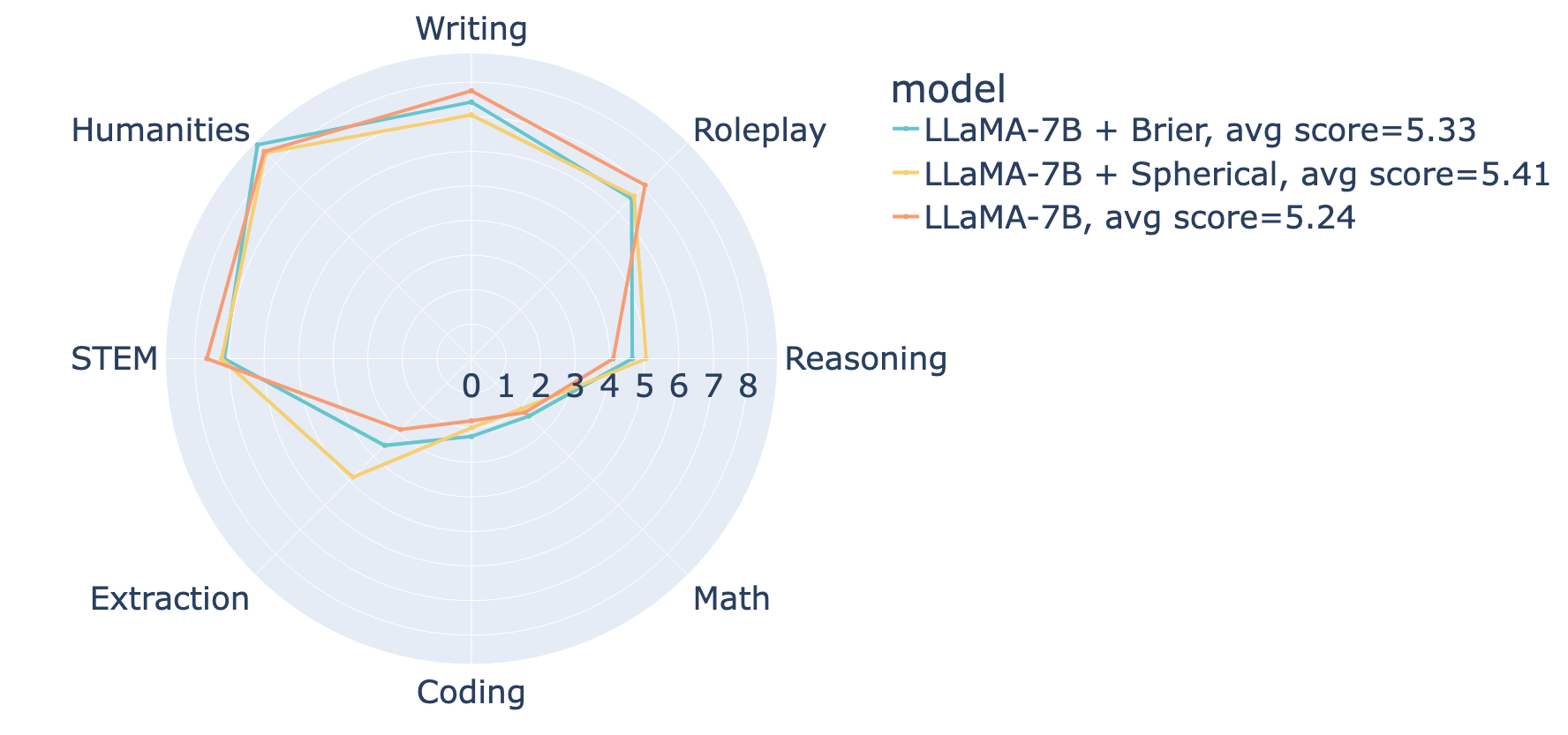}
      \vspace{-1em}
    \caption{Question answering capabilities evaluated on MT-bench, a multi-turn question set.}
    \label{fig:mtbench}
  \end{center}
  \vspace{-1em}
\end{figure}

\begin{table}[t]
\caption{ROUGE scores on CNN/DailyMail. RG-1, RG-2, RG-L stand for ROUGE-1, ROUGE-2, and ROUGE-L scores, respectively. The compared methods are based on our implementation.}
\label{table:Results_sum}
\vskip 0.1in
\begin{center}
\begin{small}
\begin{tabular}{lcccc}
\toprule
{\bf Model}  &  \bf RG-1 & \bf RG-2 & \bf RG-L \\
\midrule
Transformer  & 39.72 & 17.00 & 36.41\\
\midrule
MixCE \citep{zhang-etal-2023-mixce}& 40.16 & 17.48 & \bf 36.85 \\
TaiLr \citep{ji2023tailoring} & 39.11& 15.99 & 36.06 \\ 
Convex \citep{shao2023beyond} & 40.15 & \bf 17.67 & 36.70 \\
\midrule
Transformer + Brier &  \bf 40.20 &   17.56 &  36.78 \\
Transformer + Spherical &  \bf 40.20 &  17.55 & 36.73 \\
\bottomrule 
\end{tabular}
\end{small}
\end{center}
\vspace{-1em}
\end{table}

\begin{table}[t]
\caption{BLEU scores of Alpaca fine-tuned large language models on WMT22 test sets.}
\label{table:Results_LLAMA_MT}
\vskip 0.1in
\begin{center}
\begin{small}
\begin{tabular}{lcccc}
\toprule
{\bf Model}  & {\bf EN-DE} & {\bf DE-EN} & {\bf EN-ZH} & {\bf ZH-EN}\\
\midrule
LLaMA-7B&  25.42 & 17.93 & 13.86 & 13.17  \\
{\ \  + Brier}&  \bf{29.15} & \bf{21.09} & {15.74} & {17.75} \\
{\ \  + Spherical}&  {29.07} & {21.05} & \bf{15.87} & \bf{17.95} \\
\midrule
LLaMA-13B&  {29.35}  & 21.74  & 15.58  & 16.27  \\
{\ \  + Brier}&  {29.54} & {22.80} & \bf{17.10} & \bf{19.99}  \\
{\ \  + Spherical}&  \bf{29.82} & \bf{23.11} & {15.85} & {19.59} \\
\bottomrule
\end{tabular}
\end{small}
\end{center}
\vspace{-1em}
\end{table}

\begin{table}[t]
\caption{ROUGE scores of Alpaca fine-tuned large language models on CNN/DailyMail.}
\label{table:Results_LLAMA_CNNDM}
\vskip 0.1in
\begin{center}
\begin{small}
\begin{tabular}{lcccc}
\toprule
{\bf Model}  & {\bf RG-1} & {\bf RG-2} & {\bf RG-L} \\
\midrule
LLaMA-7B&  28.66 & 12.49 & 26.37  \\
{LLaMA-7B + Brier}& \bf{32.15} & \bf{14.76} & \bf{29.72}  \\
{LLaMA-7B + Spherical}&  {30.89} & {13.87} & {28.45}  \\
\bottomrule
\end{tabular}
\end{small}
\end{center}
\vspace{-1em}
\end{table}

\subsection{Fine-tuning with Scoring Rules}

\begin{figure*}[t]
\centering
\begin{minipage}{.33\textwidth}
  \centering
  \includegraphics[width=1.\linewidth]{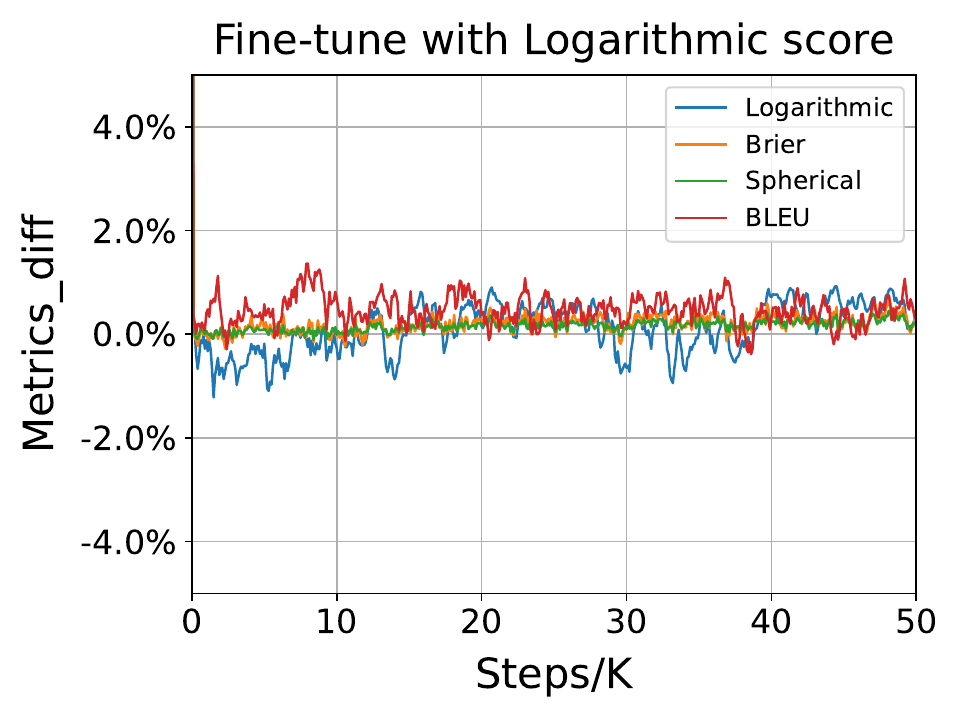}
\end{minipage}%
\begin{minipage}{.33\textwidth}
  \centering
  \includegraphics[width=1.\linewidth]{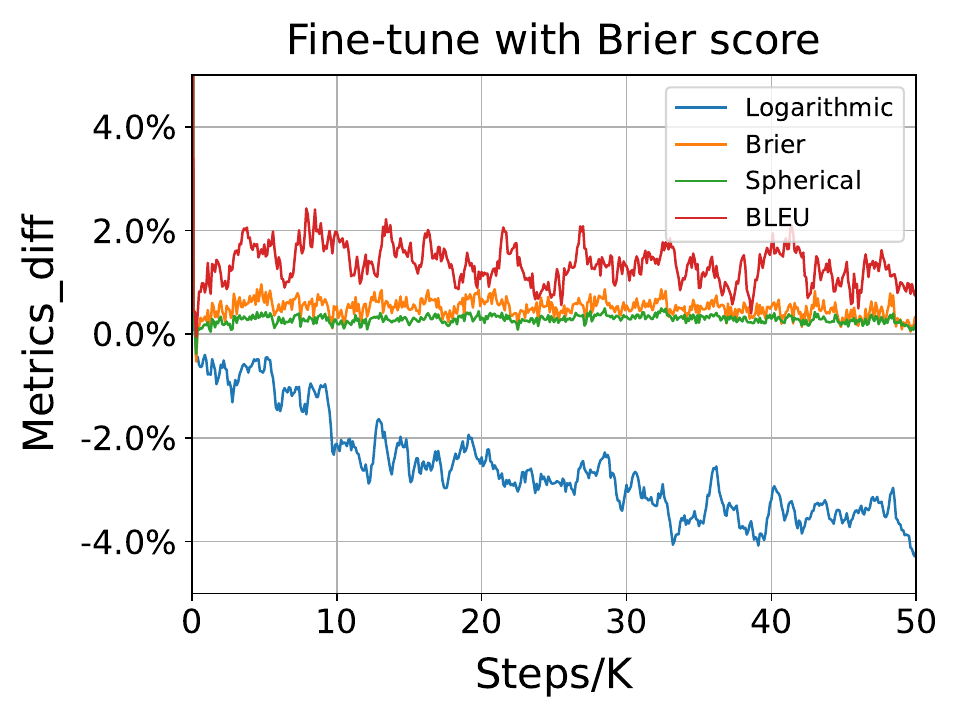}
\end{minipage}
\begin{minipage}{.33\textwidth}
  \centering
  \includegraphics[width=1.\linewidth]{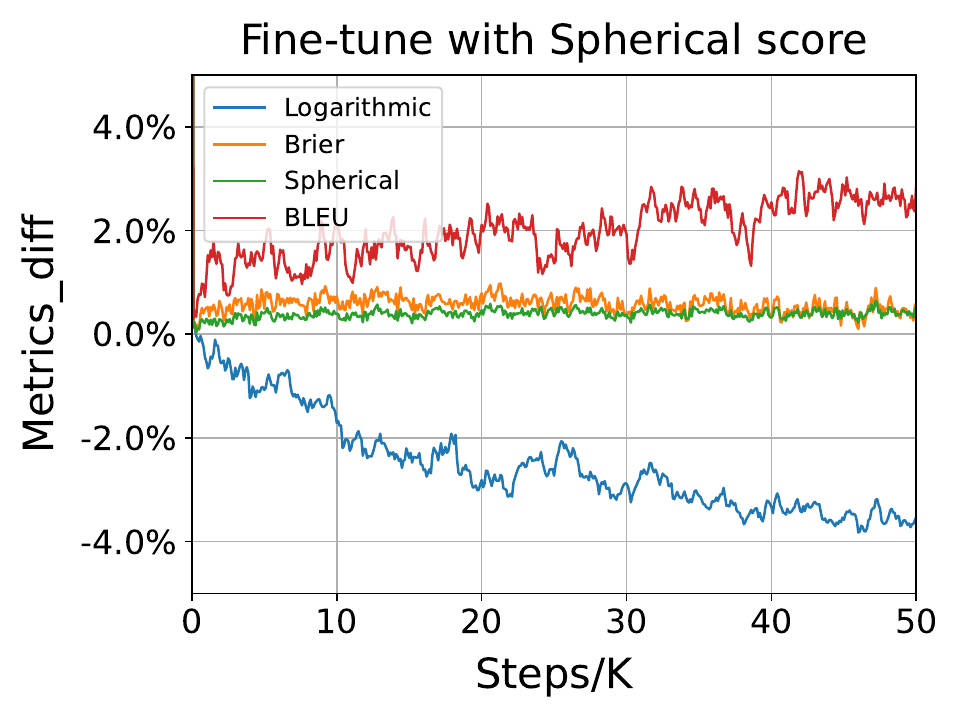}
\end{minipage}
  \caption{Performance curves on WMT14 En-De test set when fine-tuning with different scoring rules.}
  \label{fig:scores}
  \vspace{-1em}
\end{figure*}

\begin{table}[t]
\caption{BLEU scores on the WMT14 En-De test set.}
\label{table:Results_alpha_wmt14}
\vskip 0.1in
\begin{center}
\begin{small}
\setlength{\tabcolsep}{5pt}
\begin{tabular}{lccccc}
\toprule
{\bm{$\alpha$}}  &\bf 1.5 & \bf1.75 & \bf2 & \bf2.25 & \bf2.5 \\
\midrule
{$\alpha$-power}& 27.91 & 28.05 & 28.01 & 27.84 & 27.71 \\
pseudo-spherical& 28.09 & 27.91 & 28.07 & 27.92 & 27.64\\
\bottomrule
\end{tabular}
\end{small}
\end{center}
\vspace{-1em}
\end{table}

\begin{table}[t]
\caption{Average BLEU scores on WMT22 test sets.}
\label{table:Results_alpha_wmt22}
\vskip 0.1in
\begin{center}
\begin{small}
\setlength{\tabcolsep}{5pt}
\begin{tabular}{lccccc}
\toprule
{\bm{$\alpha$}}  &\bf1.5 & \bf1.75 &\bf 2 & \bf2.25 & \bf2.5 \\
\midrule
{$\alpha$-power}& 19.93 & 20.11 & 20.93 & 19.27 & 19.12 \\
pseudo-spherical& 20.42 & 20.67 & 20.98 & 20.03 & 19.52 \\
\bottomrule
\end{tabular}
\end{small}
\end{center}
\vspace{-1em}
\end{table}

\begin{table*}[t]
        \caption{BLEU scores on the TED bilingual dataset. Avg means the average BLEU. `Transformer w/ LS' represents a Transformer trained with label smoothing. `Transformer w/o LS' represents a Transformer trained without smoothing.}
        \vskip 0.1in
    \centering
    \footnotesize
    \begin{tabular}{l|ccccccccccc}
    \toprule
    \bf{X-En}   & \bf Fr & \bf Ru &\bf He &\bf  Ar & \bf   It &\bf  Nl & \bf Ro & \bf Tr  & \bf De &\bf Vi &\bf Avg \\ \midrule 
        Transformer w/o LS &	39.39&	24.81	&37.07	&31.79	&37.75	&35.86	&34.40	&25.64&	34.88	&26.48	&32.81\\
        \ \ + Brier   &	40.02&	25.43	&37.85	&32.12	&38.41	&36.32	&35.23	&26.17&	36.05	&26.81	&33.44\\
        \ \ + Spherical &	40.27	&25.49	&38.13&	32.37&	38.67&	36.85	&36.63	&26.43	&35.66&	27.02&	33.75\\
       \midrule
        Transformer w/ LS 	&40.64	&25.74	&38.48	&32.74	&38.87	&36.81	&35.77	&26.80	&36.03	&27.18	&33.91\\
        \ \ + Brier   	&40.19	&25.32	&38.36	&32.59	&38.60	&36.40	&35.40	&26.53	&35.65	&27.22	&33.63\\
        \ \ + Spherical 	&40.45	&25.87	&38.40	&32.82	&38.56	&36.68	&35.68	&26.84	&36.00	&27.34	&33.86\\
       \bottomrule
    \end{tabular}
    \label{tab:ted}
    \vspace{-0.5em}
\end{table*}

\begin{table}[t]
\caption{Average BLEU scores on TED test sets.}
\label{table:tedavg}
\vskip 0.1in
\begin{center}
\begin{small}
\begin{tabular}{lccc}
\toprule
 \bf{Score} & {$\bm{S}$} & {$\bm{S^\epsilon}$} & {$\bm{S^\epsilon_{\log}}$}\\
\midrule
Brier& 33.63 & 33.80  & 34.49\\
Spherical& 33.86 & 33.89 & 34.43\\
\bottomrule
\end{tabular}
\end{small}
\end{center}
\vspace{-1em}
\end{table}
As we have already observed, it is relatively challenging for other scoring rules to surpass the performance of the logarithmic score when training from scratch. Here, we further explore the impact of using alternative scores for fine-tuning on models trained with the logarithmic score. We fine-tune from an earlier checkpoint to ensure the total number of training steps remains unchanged. We fix all hyperparameters and only modify the loss function to correspond with the alternative scores.

Table \ref{table:Results_mt} and Table \ref{table:Results_sum} present the impact of fine-tuning on translation and summarization performance. As can be seen, even without adjusting hyperparameters for specific scores, fine-tuning with Brier score or Spherical score can still yield certain improvements on logarithmic score pre-trained models. We conjecture that such improvements stem from the complementarity between scoring rules. As different scoring rules follow unique optimization trajectories towards the same global optimum, fine-tuning with another score might aid the model in escaping its current trapped region, consequently leading to further performance improvements.

We continue to explore the effectiveness of scoring rules on a larger model scale. During the instruction tuning of LLaMA-7B and LLaMA-13B, we substitute the log-likelihood loss with loss functions associated with the Brier score and the spherical score. The translation and summarization performance are presented in Table \ref{table:Results_LLAMA_MT} and Table \ref{table:Results_LLAMA_CNNDM}, respectively. Due to memory constraints, we only assess the summarization performance of LLaMA-7B. It is surprising to see that fine-tuning with alternative scoring rules can lead to more significant performance improvements on LLMs. Particularly on LLaMA-7B, both scores exhibit an average increase of over 3 BLEU points in translation quality, and the spherical score also demonstrates an average improvement of over 3 ROUGE points in summarization performance. 

Figure \ref{fig:mtbench} displays the multi-turn question-answering capabilities of LLMs. Models fine-tuned using the Brier score and the spherical score exhibit stronger overall performance, particularly in extraction and reasoning tasks. In contrast, the model fine-tuned with the logarithmic score is better at writing, roleplay, and STEM tasks.

\subsection{Model Dynamics during Fine-tuning}

The above experiments show that fine-tuning with other scoring rules can enhance the generative capabilities of language generation models. However, it remains unclear what changes occur within the model during this process. In this section, we investigate the dynamics of the model during the fine-tuning process to better understand its impact. Specifically, on the WMT14 En-De dataset, we pre-train the Transformer using MLE loss and fine-tune it with various scoring rules. Then we track the changes of different scoring rules and also the BLEU score on the test set. Figure \ref{fig:scores} illustrates their relative changes, calculated as $\frac{S(p_{\theta},q) - S(p_{\theta_{old}},q)}{|S(p_{\theta_{old}},q)|}$.

As observed, when fine-tuning with the logarithmic score, all metrics fluctuate around their original values since the model is pre-trained with the same score. When fine-tuning with the Brier score or the spherical score, both scores show a certain improvement, accompanied by an increase in BLEU. In contrast, the logarithmic score experiences a significant drop. This interesting phenomenon implies that although different strictly proper scores share the same global optimum, their optimization trajectories might be conflicting, and these scores do not always align with the model's generative capabilities. Therefore, comprehensively considering multiple scores during training can help the model achieve stronger generative capabilities. It also suggests that when assessing language models, a more accurate evaluation could be achieved by considering multiple scores collectively, rather than relying solely on the perplexity.

\subsection{Pseudo-spherical Score and Power Score}

Previously, we explored the impact of Brier score and spherical score for training language generation models. Here, we further investigate two more general scoring rules, namely the pseudo-spherical score and the $\alpha$-power score, as described in section \ref{sec:examples}. Both scores include a parameter $\alpha$, with Brier score and spherical score being their special cases when $\alpha=2$. To examine the impact of the parameter $\alpha$, we conduct experiments on both Transformer-base and LLaMA-7B. Table \ref{table:Results_alpha_wmt14} and Table \ref{table:Results_alpha_wmt22} give the results on the WMT14 En-De test set and WMT22 test sets, respectively. Overall, a stable and superior performance is achieved at $\alpha=2$. When $\alpha>2$, the model performance typically experiences a noticeable decline. In contrast, the models can still maintain a competitive performance when $\alpha<2$.

\subsection{Effect of Score Smoothing}

Label smoothing is a commonly used regularization technique for classification networks, particularly crucial in low-resource scenarios. Therefore, we conduct experiments on the TED bilingual dataset to examine whether score smoothing could yield a similar effect. First, we train Transformer models using the smoothed and unsmoothed log-likelihood loss respectively, and then fine-tune them with the unsmoothed Brier and spherical score. The results are presented in Table \ref{tab:ted}. When not using label smoothing, fine-tuning with alternative scores brings noticeable improvements. However, for models trained with label smoothing, fine-tuning with unsmoothed scores may result in a performance decline, indicating the necessity of score smoothing.

Next, we employ score smoothing techniques to fine-tune the Transformer w/ LS. For simplicity, we only report the average BLEU score in Table \ref{table:tedavg}. The smoothed score $S^\epsilon$ results in some improvement, but the impact is relatively minor. By enhancing the smoothing term with the masked logarithmic score, ${S^\epsilon_{\log}}$ leads to a more noticeable improvement in performance, indicating that score smoothing can also serve as an effective regularization technique.

\section{Related Work}
\textbf{Strictly Proper Scoring Rules in Deep Learning.} In addition to the widely used logarithmic score, various strictly proper scoring rules have played a significant role in deep learning. The Brier score serves as a training criterion for classification networks \citep{80304,hung1996estimating,Kline,hui2021evaluation}, as well as an evaluation metric for the quality of uncertainty calibration \citep{lakshminarayanan2017simple,NEURIPS2019_8558cb40,NEURIPS2022_3915a87d}. The pseudo-spherical score offers solutions for training energy-based models \citep{NEURIPS2021_bc5fcb00} and knowledge distillation \citep{ijcai2022p441}. In the continuous space, some scoring rules present appealing generative modeling approaches. For example, the Hyv\"{a}rinen score \citep{hyvarinen2005estimation,ehm2012local} gives rise to score-based generative models \citep{NEURIPS2019_3001ef25,song2021scorebased}. The energy and kernel score \citep{gneiting2007strictly} facilitate the development of generative networks through scoring rule minimization \citep{NEURIPS2020_9873eaad,pacchiardi2021probabilistic,pacchiardi2022likelihood}. 

\vspace{5pt}
\noindent{}\textbf{Loss Functions for Language Generation.} 
Currently, the loss functions used in language generation models are primarily improved versions of cross-entropy loss. One line of research adapts the cross-entropy loss through techniques such as truncation \citep{kang-hashimoto-2020-improved} and reweighting \citep{ji2023tailoring}. Another line of research introduces an additional loss term to complement the cross-entropy loss, such as incorporating reverse cross-entropy \citep{zhang-etal-2023-mixce}, reflective likelihood loss \citep{dieng2019learning}, unlikelihood loss \citep{Welleck2020Neural}, and Gaussian prior objective \citep{Li2020Data-dependent}. \citet{stahlberg-kumar-2022-jam} transforms the multi-class word prediction problem into multiple binary classification problems, which also leads to a well-calibrated model distribution when proper scoring rules (e.g., the logarithmic score used in this work) are employed for binary classification. A recent approach \citep{shao2023beyond} involves composing the cross-entropy loss with a convex function, which results in a local but improper scoring rule that alters the shape of the model distribution to be sharper than the data distribution. Other loss functions primarily involve reinforcement learning-based reward optimization, where rewards are derived from evaluation metrics \citep{ranzato2015sequence,shen-etal-2016-minimum,shao-etal-2019-retrieving,DBLP:journals/corr/abs-2106-08122}, human feedback \citep{NEURIPS2020_1f89885d,ouyang2022training}, generative adversarial nets \citep{yu2017seqgan,yang-etal-2018-improving}, or reference demonstrations \citep{pang2021text}. To our knowledge, our work is the first attempt to train language generation models using scoring rules other than the logarithmic score.

\section{Conclusion}
This paper investigates the use of non-local strictly proper scoring rules for training language generation models, with a primary focus on the Brier score and the spherical score. Although these scores do not perform as well as the logarithmic score when training models from scratch, they demonstrate substantial improvements when fine-tuning models that have been pre-trained with the logarithmic score.

Our findings raise several intriguing questions for future research: Despite being strictly proper, different scores still exhibit considerable performance variations when training language models. Are there other scores that exhibit superior performance during pre-training or fine-tuning? What factors contribute to these performance differences, and can we develop additional metrics or properties to determine a score's suitability for training language generation models? Furthermore, beyond model training, it is worth exploring whether these scores can function as evaluation metrics, similar to Perplexity (PPL), for assessing the calibration of language generation models.

\section*{Impact Statement}
This paper presents work whose goal is to advance the field of Machine Learning. There are many potential societal consequences of our work, none which we feel must be specifically highlighted here.
\bibliography{example_paper}

\begin{thebibliography}{65}
\providecommand{\natexlab}[1]{#1}
\providecommand{\url}[1]{\texttt{#1}}
\expandafter\ifx\csname urlstyle\endcsname\relax
  \providecommand{\doi}[1]{doi: #1}\else
  \providecommand{\doi}{doi: \begingroup \urlstyle{rm}\Url}\fi

\bibitem[Bengio et~al.(2000)Bengio, Ducharme, and Vincent]{NIPS2000_728f206c}
Bengio, Y., Ducharme, R., and Vincent, P.
\newblock A neural probabilistic language model.
\newblock In Leen, T., Dietterich, T., and Tresp, V. (eds.), \emph{Advances in
  Neural Information Processing Systems}, volume~13. MIT Press, 2000.
\newblock URL
  \url{https://proceedings.neurips.cc/paper_files/paper/2000/file/728f206c2a01bf572b5940d7d9a8fa4c-Paper.pdf}.

\bibitem[Bernardo(1979)]{bernardo1979expected}
Bernardo, J.~M.
\newblock Expected information as expected utility.
\newblock \emph{the Annals of Statistics}, pp.\  686--690, 1979.

\bibitem[Brier(1950)]{brier1950verification}
Brier, G.~W.
\newblock Verification of forecasts expressed in terms of probability.
\newblock \emph{Monthly weather review}, 78\penalty0 (1):\penalty0 1--3, 1950.

\bibitem[Brown et~al.(2020)Brown, Mann, Ryder, Subbiah, Kaplan, Dhariwal,
  Neelakantan, Shyam, Sastry, Askell, et~al.]{brown2020language}
Brown, T., Mann, B., Ryder, N., Subbiah, M., Kaplan, J.~D., Dhariwal, P.,
  Neelakantan, A., Shyam, P., Sastry, G., Askell, A., et~al.
\newblock Language models are few-shot learners.
\newblock \emph{Advances in neural information processing systems},
  33:\penalty0 1877--1901, 2020.

\bibitem[Bröcker \& Smith(2007)Bröcker and
  Smith]{ScoringProbabilisticForecastsTheImportanceofBeingProper}
Bröcker, J. and Smith, L.~A.
\newblock Scoring probabilistic forecasts: The importance of being proper.
\newblock \emph{Weather and Forecasting}, 22\penalty0 (2):\penalty0 382 -- 388,
  2007.
\newblock \doi{https://doi.org/10.1175/WAF966.1}.
\newblock URL
  \url{https://journals.ametsoc.org/view/journals/wefo/22/2/waf966_1.xml}.

\bibitem[Dieng et~al.(2019)Dieng, Cho, Blei, and LeCun]{dieng2019learning}
Dieng, A.~B., Cho, K., Blei, D.~M., and LeCun, Y.
\newblock Learning with reflective likelihoods, 2019.
\newblock URL \url{https://openreview.net/forum?id=SJlh2jR9FX}.

\bibitem[Ehm \& Gneiting(2012)Ehm and Gneiting]{ehm2012local}
Ehm, W. and Gneiting, T.
\newblock Local proper scoring rules of order two.
\newblock \emph{The Annals of Statistics}, 40\penalty0 (1):\penalty0 609--637,
  2012.

\bibitem[Gneiting \& Raftery(2007)Gneiting and Raftery]{gneiting2007strictly}
Gneiting, T. and Raftery, A.~E.
\newblock Strictly proper scoring rules, prediction, and estimation.
\newblock \emph{Journal of the American statistical Association}, 102\penalty0
  (477):\penalty0 359--378, 2007.

\bibitem[Good(1952)]{good1952rational}
Good, I.~J.
\newblock Rational decisions.
\newblock \emph{Journal of the Royal Statistical Society: Series B
  (Methodological)}, 14\penalty0 (1):\penalty0 107--114, 1952.

\bibitem[Gritsenko et~al.(2020)Gritsenko, Salimans, van~den Berg, Snoek, and
  Kalchbrenner]{NEURIPS2020_9873eaad}
Gritsenko, A., Salimans, T., van~den Berg, R., Snoek, J., and Kalchbrenner, N.
\newblock A spectral energy distance for parallel speech synthesis.
\newblock In Larochelle, H., Ranzato, M., Hadsell, R., Balcan, M., and Lin, H.
  (eds.), \emph{Advances in Neural Information Processing Systems}, volume~33,
  pp.\  13062--13072. Curran Associates, Inc., 2020.
\newblock URL
  \url{https://proceedings.neurips.cc/paper_files/paper/2020/file/9873eaad153c6c960616c89e54fe155a-Paper.pdf}.

\bibitem[Gruber \& Buettner(2022)Gruber and Buettner]{NEURIPS2022_3915a87d}
Gruber, S. and Buettner, F.
\newblock Better uncertainty calibration via proper scores for classification
  and beyond.
\newblock In Koyejo, S., Mohamed, S., Agarwal, A., Belgrave, D., Cho, K., and
  Oh, A. (eds.), \emph{Advances in Neural Information Processing Systems},
  volume~35, pp.\  8618--8632. Curran Associates, Inc., 2022.
\newblock URL
  \url{https://proceedings.neurips.cc/paper_files/paper/2022/file/3915a87ddac8e8c2f23dbabbcee6eec9-Paper-Conference.pdf}.

\bibitem[Hermann et~al.(2015)Hermann, Kocisky, Grefenstette, Espeholt, Kay,
  Suleyman, and Blunsom]{NIPS2015_afdec700}
Hermann, K.~M., Kocisky, T., Grefenstette, E., Espeholt, L., Kay, W., Suleyman,
  M., and Blunsom, P.
\newblock Teaching machines to read and comprehend.
\newblock In Cortes, C., Lawrence, N., Lee, D., Sugiyama, M., and Garnett, R.
  (eds.), \emph{Advances in Neural Information Processing Systems}, volume~28.
  Curran Associates, Inc., 2015.
\newblock URL
  \url{https://proceedings.neurips.cc/paper_files/paper/2015/file/afdec7005cc9f14302cd0474fd0f3c96-Paper.pdf}.

\bibitem[Hui \& Belkin(2021)Hui and Belkin]{hui2021evaluation}
Hui, L. and Belkin, M.
\newblock Evaluation of neural architectures trained with square loss vs
  cross-entropy in classification tasks.
\newblock In \emph{International Conference on Learning Representations}, 2021.
\newblock URL \url{https://openreview.net/forum?id=hsFN92eQEla}.

\bibitem[Hung et~al.(1996)Hung, Hu, Shanker, and Patuwo]{hung1996estimating}
Hung, M., Hu, M., Shanker, M., and Patuwo, B.
\newblock Estimating posterior probabilities in classification problems with
  neural networks.
\newblock \emph{International Journal of Computational Intelligence and
  Organizations}, 1\penalty0 (1):\penalty0 49--60, 1996.

\bibitem[Hyv{\"a}rinen \& Dayan(2005)Hyv{\"a}rinen and
  Dayan]{hyvarinen2005estimation}
Hyv{\"a}rinen, A. and Dayan, P.
\newblock Estimation of non-normalized statistical models by score matching.
\newblock \emph{Journal of Machine Learning Research}, 6\penalty0 (4), 2005.

\bibitem[Ji et~al.(2023)Ji, Ke, Hu, Zhang, and Huang]{ji2023tailoring}
Ji, H., Ke, P., Hu, Z., Zhang, R., and Huang, M.
\newblock Tailoring language generation models under total variation distance.
\newblock In \emph{The Eleventh International Conference on Learning
  Representations}, 2023.
\newblock URL \url{https://openreview.net/forum?id=VELL0PlWfc}.

\bibitem[Jiao et~al.(2023)Jiao, tse Huang, Wang, Wang, Shi, and
  Tu]{jiao2023parrot}
Jiao, W., tse Huang, J., Wang, W., Wang, X., Shi, S., and Tu, Z.
\newblock Parrot: Translating during chat using large language models.
\newblock \emph{arXiv preprint arXiv:2304.02426}, 2023.

\bibitem[Kang \& Hashimoto(2020)Kang and
  Hashimoto]{kang-hashimoto-2020-improved}
Kang, D. and Hashimoto, T.~B.
\newblock Improved natural language generation via loss truncation.
\newblock In Jurafsky, D., Chai, J., Schluter, N., and Tetreault, J. (eds.),
  \emph{Proceedings of the 58th Annual Meeting of the Association for
  Computational Linguistics}, pp.\  718--731, Online, July 2020. Association
  for Computational Linguistics.
\newblock \doi{10.18653/v1/2020.acl-main.66}.
\newblock URL \url{https://aclanthology.org/2020.acl-main.66}.

\bibitem[Kline \& Berardi(2005)Kline and Berardi]{Kline}
Kline, D. and Berardi, V.
\newblock Revisiting squared-error and cross-entropy functions for training
  neural network classifiers.
\newblock \emph{Neural Computing and Applications}, 14:\penalty0 310--318, 12
  2005.
\newblock \doi{10.1007/s00521-005-0467-y}.

\bibitem[Lakshminarayanan et~al.(2017)Lakshminarayanan, Pritzel, and
  Blundell]{lakshminarayanan2017simple}
Lakshminarayanan, B., Pritzel, A., and Blundell, C.
\newblock Simple and scalable predictive uncertainty estimation using deep
  ensembles.
\newblock \emph{Advances in neural information processing systems}, 30, 2017.

\bibitem[Lee \& Lee(2022)Lee and Lee]{ijcai2022p441}
Lee, K. and Lee, H.
\newblock Pseudo-spherical knowledge distillation.
\newblock In Raedt, L.~D. (ed.), \emph{Proceedings of the Thirty-First
  International Joint Conference on Artificial Intelligence, {IJCAI-22}}, pp.\
  3178--3184. International Joint Conferences on Artificial Intelligence
  Organization, 7 2022.
\newblock \doi{10.24963/ijcai.2022/441}.
\newblock URL \url{https://doi.org/10.24963/ijcai.2022/441}.
\newblock Main Track.

\bibitem[Li et~al.(2020)Li, Wang, Chen, Utiyama, Sumita, Zhang, and
  Zhao]{Li2020Data-dependent}
Li, Z., Wang, R., Chen, K., Utiyama, M., Sumita, E., Zhang, Z., and Zhao, H.
\newblock Data-dependent gaussian prior objective for language generation.
\newblock In \emph{International Conference on Learning Representations}, 2020.
\newblock URL \url{https://openreview.net/forum?id=S1efxTVYDr}.

\bibitem[Lin(2004)]{lin-2004-rouge}
Lin, C.-Y.
\newblock {ROUGE}: A package for automatic evaluation of summaries.
\newblock In \emph{Text Summarization Branches Out}, pp.\  74--81, Barcelona,
  Spain, July 2004. Association for Computational Linguistics.
\newblock URL \url{https://aclanthology.org/W04-1013}.

\bibitem[Liu et~al.(2021)Liu, Yan, Gong, Qi, Zhang, Jiao, Chen, Fu, Shou, Gong,
  Wang, Chen, Jiang, Lv, Zhang, Wu, Zhou, and
  Duan]{DBLP:conf/acl/LiuYGQZJCFSGWCJ21}
Liu, D., Yan, Y., Gong, Y., Qi, W., Zhang, H., Jiao, J., Chen, W., Fu, J.,
  Shou, L., Gong, M., Wang, P., Chen, J., Jiang, D., Lv, J., Zhang, R., Wu, W.,
  Zhou, M., and Duan, N.
\newblock {GLGE:} {A} new general language generation evaluation benchmark.
\newblock In Zong, C., Xia, F., Li, W., and Navigli, R. (eds.), \emph{Findings
  of the Association for Computational Linguistics: {ACL/IJCNLP} 2021, Online
  Event, August 1-6, 2021}, volume {ACL/IJCNLP} 2021 of \emph{Findings of
  {ACL}}, pp.\  408--420. Association for Computational Linguistics, 2021.
\newblock \doi{10.18653/v1/2021.findings-acl.36}.
\newblock URL \url{https://doi.org/10.18653/v1/2021.findings-acl.36}.

\bibitem[Liu et~al.(2023)Liu, Zeng, Meng, and Zhou]{liu2023instruction}
Liu, Y., Zeng, X., Meng, F., and Zhou, J.
\newblock Instruction position matters in sequence generation with large
  language models.
\newblock \emph{arXiv preprint arXiv:2308.12097}, 2023.

\bibitem[Martins \& Astudillo(2016)Martins and Astudillo]{pmlr-v48-martins16}
Martins, A. and Astudillo, R.
\newblock From softmax to sparsemax: A sparse model of attention and
  multi-label classification.
\newblock In Balcan, M.~F. and Weinberger, K.~Q. (eds.), \emph{Proceedings of
  The 33rd International Conference on Machine Learning}, volume~48 of
  \emph{Proceedings of Machine Learning Research}, pp.\  1614--1623, New York,
  New York, USA, 20--22 Jun 2016. PMLR.
\newblock URL \url{https://proceedings.mlr.press/v48/martins16.html}.

\bibitem[Martins et~al.(2020)Martins, Marinho, and
  Martins]{martins-etal-2020-sparse}
Martins, P.~H., Marinho, Z., and Martins, A. F.~T.
\newblock Sparse text generation.
\newblock In Webber, B., Cohn, T., He, Y., and Liu, Y. (eds.),
  \emph{Proceedings of the 2020 Conference on Empirical Methods in Natural
  Language Processing (EMNLP)}, pp.\  4252--4273, Online, November 2020.
  Association for Computational Linguistics.
\newblock \doi{10.18653/v1/2020.emnlp-main.348}.
\newblock URL \url{https://aclanthology.org/2020.emnlp-main.348}.

\bibitem[Mikolov et~al.(2010)Mikolov, Karafi{\'a}t, Burget, {\v C}ernock{\'y},
  and Khudanpur]{Mikolov2010RecurrentNN}
Mikolov, T., Karafi{\'a}t, M., Burget, L., {\v C}ernock{\'y}, J.~H., and
  Khudanpur, S.
\newblock Recurrent neural network based language model.
\newblock In \emph{Interspeech}, 2010.
\newblock URL \url{https://api.semanticscholar.org/CorpusID:17048224}.

\bibitem[Myung(2003)]{myung2003tutorial}
Myung, I.~J.
\newblock Tutorial on maximum likelihood estimation.
\newblock \emph{Journal of mathematical Psychology}, 47\penalty0 (1):\penalty0
  90--100, 2003.

\bibitem[Ouyang et~al.(2022)Ouyang, Wu, Jiang, Almeida, Wainwright, Mishkin,
  Zhang, Agarwal, Slama, Gray, Schulman, Hilton, Kelton, Miller, Simens,
  Askell, Welinder, Christiano, Leike, and Lowe]{ouyang2022training}
Ouyang, L., Wu, J., Jiang, X., Almeida, D., Wainwright, C., Mishkin, P., Zhang,
  C., Agarwal, S., Slama, K., Gray, A., Schulman, J., Hilton, J., Kelton, F.,
  Miller, L., Simens, M., Askell, A., Welinder, P., Christiano, P., Leike, J.,
  and Lowe, R.
\newblock Training language models to follow instructions with human feedback.
\newblock In Oh, A.~H., Agarwal, A., Belgrave, D., and Cho, K. (eds.),
  \emph{Advances in Neural Information Processing Systems}, 2022.
\newblock URL \url{https://openreview.net/forum?id=TG8KACxEON}.

\bibitem[Ovadia et~al.(2019)Ovadia, Fertig, Ren, Nado, Sculley, Nowozin,
  Dillon, Lakshminarayanan, and Snoek]{NEURIPS2019_8558cb40}
Ovadia, Y., Fertig, E., Ren, J., Nado, Z., Sculley, D., Nowozin, S., Dillon,
  J., Lakshminarayanan, B., and Snoek, J.
\newblock Can you trust your model\textquotesingle s uncertainty? evaluating
  predictive uncertainty under dataset shift.
\newblock In Wallach, H., Larochelle, H., Beygelzimer, A., d\textquotesingle
  Alch\'{e}-Buc, F., Fox, E., and Garnett, R. (eds.), \emph{Advances in Neural
  Information Processing Systems}, volume~32. Curran Associates, Inc., 2019.
\newblock URL
  \url{https://proceedings.neurips.cc/paper_files/paper/2019/file/8558cb408c1d76621371888657d2eb1d-Paper.pdf}.

\bibitem[Pacchiardi \& Dutta(2022)Pacchiardi and
  Dutta]{pacchiardi2022likelihood}
Pacchiardi, L. and Dutta, R.
\newblock Likelihood-free inference with generative neural networks via scoring
  rule minimization.
\newblock \emph{arXiv preprint arXiv:2205.15784}, 2022.

\bibitem[Pacchiardi et~al.(2021)Pacchiardi, Adewoyin, Dueben, and
  Dutta]{pacchiardi2021probabilistic}
Pacchiardi, L., Adewoyin, R., Dueben, P., and Dutta, R.
\newblock Probabilistic forecasting with generative networks via scoring rule
  minimization.
\newblock \emph{arXiv preprint arXiv:2112.08217}, 2021.

\bibitem[Pang \& He(2021)Pang and He]{pang2021text}
Pang, R.~Y. and He, H.
\newblock Text generation by learning from demonstrations.
\newblock In \emph{International Conference on Learning Representations}, 2021.
\newblock URL \url{https://openreview.net/forum?id=RovX-uQ1Hua}.

\bibitem[Papineni et~al.(2002)Papineni, Roukos, Ward, and
  Zhu]{papineni-etal-2002-bleu}
Papineni, K., Roukos, S., Ward, T., and Zhu, W.-J.
\newblock {B}leu: a method for automatic evaluation of machine translation.
\newblock In \emph{Proceedings of the 40th Annual Meeting of the Association
  for Computational Linguistics}, pp.\  311--318, Philadelphia, Pennsylvania,
  USA, July 2002. Association for Computational Linguistics.
\newblock \doi{10.3115/1073083.1073135}.
\newblock URL \url{https://aclanthology.org/P02-1040}.

\bibitem[Peters et~al.(2019)Peters, Niculae, and
  Martins]{peters-etal-2019-sparse}
Peters, B., Niculae, V., and Martins, A. F.~T.
\newblock Sparse sequence-to-sequence models.
\newblock In Korhonen, A., Traum, D., and M{\`a}rquez, L. (eds.),
  \emph{Proceedings of the 57th Annual Meeting of the Association for
  Computational Linguistics}, pp.\  1504--1519, Florence, Italy, July 2019.
  Association for Computational Linguistics.
\newblock \doi{10.18653/v1/P19-1146}.
\newblock URL \url{https://aclanthology.org/P19-1146}.

\bibitem[Radford et~al.(2018)Radford, Narasimhan, Salimans, Sutskever,
  et~al.]{radford2018improving}
Radford, A., Narasimhan, K., Salimans, T., Sutskever, I., et~al.
\newblock Improving language understanding by generative pre-training.
\newblock 2018.

\bibitem[Ranzato et~al.(2016)Ranzato, Chopra, Auli, and
  Zaremba]{ranzato2015sequence}
Ranzato, M., Chopra, S., Auli, M., and Zaremba, W.
\newblock Sequence level training with recurrent neural networks.
\newblock In Bengio, Y. and LeCun, Y. (eds.), \emph{4th International
  Conference on Learning Representations, {ICLR} 2016, San Juan, Puerto Rico,
  May 2-4, 2016, Conference Track Proceedings}, 2016.
\newblock URL \url{http://arxiv.org/abs/1511.06732}.

\bibitem[Roby(1965)]{roby1965belief}
Roby, T.~B.
\newblock Belief states: A preliminary empirical study.
\newblock \emph{Behavioral Sci}, 10\penalty0 (3):\penalty0 255--270, 1965.

\bibitem[See et~al.(2017)See, Liu, and Manning]{see-etal-2017-get}
See, A., Liu, P.~J., and Manning, C.~D.
\newblock Get to the point: Summarization with pointer-generator networks.
\newblock In \emph{Proceedings of the 55th Annual Meeting of the Association
  for Computational Linguistics (Volume 1: Long Papers)}, pp.\  1073--1083,
  Vancouver, Canada, July 2017. Association for Computational Linguistics.
\newblock \doi{10.18653/v1/P17-1099}.
\newblock URL \url{https://aclanthology.org/P17-1099}.

\bibitem[Selten(1998)]{selten1998axiomatic}
Selten, R.
\newblock Axiomatic characterization of the quadratic scoring rule.
\newblock \emph{Experimental Economics}, 1:\penalty0 43--61, 1998.

\bibitem[Sennrich et~al.(2016)Sennrich, Haddow, and
  Birch]{DBLP:conf/acl/SennrichHB16a}
Sennrich, R., Haddow, B., and Birch, A.
\newblock Neural machine translation of rare words with subword units.
\newblock In \emph{Proceedings of the 54th Annual Meeting of the Association
  for Computational Linguistics, {ACL} 2016, August 7-12, 2016, Berlin,
  Germany, Volume 1: Long Papers}. The Association for Computer Linguistics,
  2016.
\newblock \doi{10.18653/v1/p16-1162}.
\newblock URL \url{https://doi.org/10.18653/v1/p16-1162}.

\bibitem[Shao et~al.(2019)Shao, Feng, Zhang, Meng, Chen, and
  Zhou]{shao-etal-2019-retrieving}
Shao, C., Feng, Y., Zhang, J., Meng, F., Chen, X., and Zhou, J.
\newblock Retrieving sequential information for non-autoregressive neural
  machine translation.
\newblock In \emph{Proceedings of the 57th Annual Meeting of the Association
  for Computational Linguistics}, pp.\  3013--3024, Florence, Italy, July 2019.
  Association for Computational Linguistics.
\newblock \doi{10.18653/v1/P19-1288}.
\newblock URL \url{https://www.aclweb.org/anthology/P19-1288}.

\bibitem[Shao et~al.(2021)Shao, Feng, Zhang, Meng, and
  Zhou]{DBLP:journals/corr/abs-2106-08122}
Shao, C., Feng, Y., Zhang, J., Meng, F., and Zhou, J.
\newblock {Sequence-Level Training for Non-Autoregressive Neural Machine
  Translation}.
\newblock \emph{Computational Linguistics}, pp.\  1--35, 10 2021.
\newblock ISSN 0891-2017.
\newblock \doi{10.1162/coli_a_00421}.
\newblock URL \url{https://doi.org/10.1162/coli\_a\_00421}.

\bibitem[Shao et~al.(2023)Shao, Ma, Zhang, and Feng]{shao2023beyond}
Shao, C., Ma, Z., Zhang, M., and Feng, Y.
\newblock Beyond mle: Convex learning for text generation.
\newblock In \emph{Thirty-seventh Conference on Neural Information Processing
  Systems}, 2023.

\bibitem[Shen et~al.(2016)Shen, Cheng, He, He, Wu, Sun, and
  Liu]{shen-etal-2016-minimum}
Shen, S., Cheng, Y., He, Z., He, W., Wu, H., Sun, M., and Liu, Y.
\newblock Minimum risk training for neural machine translation.
\newblock In \emph{Proceedings of the 54th Annual Meeting of the Association
  for Computational Linguistics (Volume 1: Long Papers)}, pp.\  1683--1692,
  Berlin, Germany, August 2016. Association for Computational Linguistics.
\newblock \doi{10.18653/v1/P16-1159}.
\newblock URL \url{https://aclanthology.org/P16-1159}.

\bibitem[Shoemaker(1991)]{80304}
Shoemaker, P.
\newblock A note on least-squares learning procedures and classification by
  neural network models.
\newblock \emph{IEEE Transactions on Neural Networks}, 2\penalty0 (1):\penalty0
  158--160, 1991.
\newblock \doi{10.1109/72.80304}.

\bibitem[Shuford~Jr et~al.(1966)Shuford~Jr, Albert, and
  Edward~Massengill]{shuford1966admissible}
Shuford~Jr, E.~H., Albert, A., and Edward~Massengill, H.
\newblock Admissible probability measurement procedures.
\newblock \emph{Psychometrika}, 31\penalty0 (2):\penalty0 125--145, 1966.

\bibitem[Song \& Ermon(2019)Song and Ermon]{NEURIPS2019_3001ef25}
Song, Y. and Ermon, S.
\newblock Generative modeling by estimating gradients of the data distribution.
\newblock In Wallach, H., Larochelle, H., Beygelzimer, A., d\textquotesingle
  Alch\'{e}-Buc, F., Fox, E., and Garnett, R. (eds.), \emph{Advances in Neural
  Information Processing Systems}, volume~32. Curran Associates, Inc., 2019.
\newblock URL
  \url{https://proceedings.neurips.cc/paper_files/paper/2019/file/3001ef257407d5a371a96dcd947c7d93-Paper.pdf}.

\bibitem[Song et~al.(2021)Song, Sohl-Dickstein, Kingma, Kumar, Ermon, and
  Poole]{song2021scorebased}
Song, Y., Sohl-Dickstein, J., Kingma, D.~P., Kumar, A., Ermon, S., and Poole,
  B.
\newblock Score-based generative modeling through stochastic differential
  equations.
\newblock In \emph{International Conference on Learning Representations}, 2021.
\newblock URL \url{https://openreview.net/forum?id=PxTIG12RRHS}.

\bibitem[Stahlberg \& Kumar(2022)Stahlberg and Kumar]{stahlberg-kumar-2022-jam}
Stahlberg, F. and Kumar, S.
\newblock Jam or cream first? modeling ambiguity in neural machine translation
  with {SCONES}.
\newblock In Carpuat, M., de~Marneffe, M.-C., and Meza~Ruiz, I.~V. (eds.),
  \emph{Proceedings of the 2022 Conference of the North American Chapter of the
  Association for Computational Linguistics: Human Language Technologies}, pp.\
   4950--4961, Seattle, United States, July 2022. Association for Computational
  Linguistics.
\newblock \doi{10.18653/v1/2022.naacl-main.365}.
\newblock URL \url{https://aclanthology.org/2022.naacl-main.365}.

\bibitem[Stiennon et~al.(2020)Stiennon, Ouyang, Wu, Ziegler, Lowe, Voss,
  Radford, Amodei, and Christiano]{NEURIPS2020_1f89885d}
Stiennon, N., Ouyang, L., Wu, J., Ziegler, D., Lowe, R., Voss, C., Radford, A.,
  Amodei, D., and Christiano, P.~F.
\newblock Learning to summarize with human feedback.
\newblock In Larochelle, H., Ranzato, M., Hadsell, R., Balcan, M., and Lin, H.
  (eds.), \emph{Advances in Neural Information Processing Systems}, volume~33,
  pp.\  3008--3021. Curran Associates, Inc., 2020.
\newblock URL
  \url{https://proceedings.neurips.cc/paper_files/paper/2020/file/1f89885d556929e98d3ef9b86448f951-Paper.pdf}.

\bibitem[Szegedy et~al.(2016)Szegedy, Vanhoucke, Ioffe, Shlens, and
  Wojna]{szegedy2016rethinking}
Szegedy, C., Vanhoucke, V., Ioffe, S., Shlens, J., and Wojna, Z.
\newblock Rethinking the inception architecture for computer vision.
\newblock In \emph{Proceedings of the IEEE conference on computer vision and
  pattern recognition}, pp.\  2818--2826, 2016.

\bibitem[Taori et~al.(2023)Taori, Gulrajani, Zhang, Dubois, Li, Guestrin,
  Liang, and Hashimoto]{alpaca}
Taori, R., Gulrajani, I., Zhang, T., Dubois, Y., Li, X., Guestrin, C., Liang,
  P., and Hashimoto, T.~B.
\newblock Stanford alpaca: An instruction-following llama model.
\newblock \url{https://github.com/tatsu-lab/stanford_alpaca}, 2023.

\bibitem[Touvron et~al.(2023)Touvron, Lavril, Izacard, Martinet, Lachaux,
  Lacroix, Rozière, Goyal, Hambro, Azhar, Rodriguez, Joulin, Grave, and
  Lample]{touvron2023llama}
Touvron, H., Lavril, T., Izacard, G., Martinet, X., Lachaux, M.-A., Lacroix,
  T., Rozière, B., Goyal, N., Hambro, E., Azhar, F., Rodriguez, A., Joulin,
  A., Grave, E., and Lample, G.
\newblock Llama: Open and efficient foundation language models.
\newblock \emph{arXiv preprint arXiv:2302.13971}, 2023.

\bibitem[Vaswani et~al.(2017)Vaswani, Shazeer, Parmar, Uszkoreit, Jones, Gomez,
  Kaiser, and Polosukhin]{DBLP:conf/nips/VaswaniSPUJGKP17}
Vaswani, A., Shazeer, N., Parmar, N., Uszkoreit, J., Jones, L., Gomez, A.~N.,
  Kaiser, L.~u., and Polosukhin, I.
\newblock Attention is all you need.
\newblock In Guyon, I., Luxburg, U.~V., Bengio, S., Wallach, H., Fergus, R.,
  Vishwanathan, S., and Garnett, R. (eds.), \emph{Advances in Neural
  Information Processing Systems}, volume~30. Curran Associates, Inc., 2017.
\newblock URL
  \url{https://proceedings.neurips.cc/paper_files/paper/2017/file/3f5ee243547dee91fbd053c1c4a845aa-Paper.pdf}.

\bibitem[Wang et~al.(2022)Wang, Kordi, Mishra, Liu, Smith, Khashabi, and
  Hajishirzi]{selfinstruct}
Wang, Y., Kordi, Y., Mishra, S., Liu, A., Smith, N.~A., Khashabi, D., and
  Hajishirzi, H.
\newblock Self-instruct: Aligning language model with self generated
  instructions.
\newblock \emph{arXiv preprint arXiv:2212.10560}, 2022.

\bibitem[Welleck et~al.(2020)Welleck, Kulikov, Roller, Dinan, Cho, and
  Weston]{Welleck2020Neural}
Welleck, S., Kulikov, I., Roller, S., Dinan, E., Cho, K., and Weston, J.
\newblock Neural text generation with unlikelihood training.
\newblock In \emph{International Conference on Learning Representations}, 2020.
\newblock URL \url{https://openreview.net/forum?id=SJeYe0NtvH}.

\bibitem[Xu et~al.(2021)Xu, Zhou, Gan, Zheng, and Li]{xu-etal-2021-vocabulary}
Xu, J., Zhou, H., Gan, C., Zheng, Z., and Li, L.
\newblock Vocabulary learning via optimal transport for neural machine
  translation.
\newblock In \emph{Proceedings of the 59th Annual Meeting of the Association
  for Computational Linguistics and the 11th International Joint Conference on
  Natural Language Processing (Volume 1: Long Papers)}, pp.\  7361--7373,
  Online, August 2021. Association for Computational Linguistics.
\newblock \doi{10.18653/v1/2021.acl-long.571}.
\newblock URL \url{https://aclanthology.org/2021.acl-long.571}.

\bibitem[Yang et~al.(2018)Yang, Chen, Wang, and Xu]{yang-etal-2018-improving}
Yang, Z., Chen, W., Wang, F., and Xu, B.
\newblock Improving neural machine translation with conditional sequence
  generative adversarial nets.
\newblock In Walker, M., Ji, H., and Stent, A. (eds.), \emph{Proceedings of the
  2018 Conference of the North {A}merican Chapter of the Association for
  Computational Linguistics: Human Language Technologies, Volume 1 (Long
  Papers)}, pp.\  1346--1355, New Orleans, Louisiana, June 2018. Association
  for Computational Linguistics.
\newblock \doi{10.18653/v1/N18-1122}.
\newblock URL \url{https://aclanthology.org/N18-1122}.

\bibitem[Yu et~al.(2017)Yu, Zhang, Wang, and Yu]{yu2017seqgan}
Yu, L., Zhang, W., Wang, J., and Yu, Y.
\newblock Seqgan: Sequence generative adversarial nets with policy gradient.
\newblock In \emph{Proceedings of the Thirty-First AAAI Conference on
  Artificial Intelligence}, AAAI'17, pp.\  2852--2858. AAAI Press, 2017.

\bibitem[Yu et~al.(2021)Yu, Song, Song, and Ermon]{NEURIPS2021_bc5fcb00}
Yu, L., Song, J., Song, Y., and Ermon, S.
\newblock Pseudo-spherical contrastive divergence.
\newblock In Ranzato, M., Beygelzimer, A., Dauphin, Y., Liang, P., and Vaughan,
  J.~W. (eds.), \emph{Advances in Neural Information Processing Systems},
  volume~34, pp.\  22348--22362. Curran Associates, Inc., 2021.
\newblock URL
  \url{https://proceedings.neurips.cc/paper_files/paper/2021/file/bc5fcb0018cecacba559dc512740091b-Paper.pdf}.

\bibitem[Zeng et~al.(2023)Zeng, Meng, Yin, and Zhou]{zeng2023tim}
Zeng, J., Meng, F., Yin, Y., and Zhou, J.
\newblock Tim: Teaching large language models to translate with comparison.
\newblock \emph{arXiv preprint arXiv:2307.04408}, 2023.

\bibitem[Zhang et~al.(2023{\natexlab{a}})Zhang, Fang, Zhang, Ma, Zhou, Huang,
  Bu, Gui, Chen, Chen, and Feng]{bayling}
Zhang, S., Fang, Q., Zhang, Z., Ma, Z., Zhou, Y., Huang, L., Bu, M., Gui, S.,
  Chen, Y., Chen, X., and Feng, Y.
\newblock Bayling: Bridging cross-lingual alignment and instruction following
  through interactive translation for large language models.
\newblock \emph{arXiv preprint arXiv:2306.10968}, 2023{\natexlab{a}}.

\bibitem[Zhang et~al.(2023{\natexlab{b}})Zhang, Wu, Irsoy, Lu, Bansal, Dredze,
  and Rosenberg]{zhang-etal-2023-mixce}
Zhang, S., Wu, S., Irsoy, O., Lu, S., Bansal, M., Dredze, M., and Rosenberg, D.
\newblock {M}ix{CE}: Training autoregressive language models by mixing forward
  and reverse cross-entropies.
\newblock In Rogers, A., Boyd-Graber, J., and Okazaki, N. (eds.),
  \emph{Proceedings of the 61st Annual Meeting of the Association for
  Computational Linguistics (Volume 1: Long Papers)}, pp.\  9027--9050,
  Toronto, Canada, July 2023{\natexlab{b}}. Association for Computational
  Linguistics.
\newblock \doi{10.18653/v1/2023.acl-long.502}.
\newblock URL \url{https://aclanthology.org/2023.acl-long.502}.

\end{thebibliography}
\bibliographystyle{icml2024}

\newpage
\appendix
\onecolumn

\section{$\alpha$-power Score and $\alpha$-entmax Loss}
We are grateful for the very insightful comments provided by Reviewer gaCL, which motivated us to investigate the connection between the $\alpha$-power score and $\alpha$-entmax loss.

Softmax has a limitation in that it cannot produce probabilities exactly equal to zero. To generate sparse probability distributions, methods such as sparsemax \citep{pmlr-v48-martins16} and $\alpha$-entmax \citep{peters-etal-2019-sparse,martins-etal-2020-sparse} have been proposed, where sparsemax is a special case of $\alpha$-entmax with $\alpha=2$. Given the probability space $\triangle^d = \{p \in \mathbb{R}^d \colon p \geq 0, \|p\|_1 = 1\}$, $\alpha$-entmax is a transformation $\mathbb{R}^d \rightarrow \triangle^d$, defined as:
\begin{equation}
\label{eq:argmax}
\alpha\text{-entmax}({z}) = \arg\max_{p \in \triangle^d}p^\top z + H^\top_{\alpha}(p),
\end{equation}
where $H^\top_{\alpha}(p)$ is a family of entropies parametrized by a scalar $\alpha \geq 1$, known as Tsallis $\alpha$-entropies: 
\begin{equation}
H^\top_{\alpha}(p)= \begin{cases}
      \frac{1}{\alpha(\alpha-1)}\sum_j {(p_j-p_j^{\alpha})}, & \alpha > 1 \\
      -\sum_j p_j \log p_j, & \alpha=1 \\
      \end{cases}.
\end{equation}
The associated loss function is called $\alpha$-entmax loss:
\begin{equation}
\label{eq:entmaxloss}
\mathcal{L}_{\alpha}(z,x)= (p-e_x)^\top z + H^\top_{\alpha}(p),
\end{equation}
where  $p=\alpha\text{-entmax}({z})$ and $e_x$ is the one-hot vector corresponding to the ground truth word $x$.
We will show that under certain conditions (i.e., the probability of ground truth word $p_x > 0$, $\alpha>1$), the $\alpha$-entmax loss is equivalent to the following token-level loss based on $\alpha$-power score:
\begin{equation}
\mathcal{L}_{\alpha\text{-power}}(p,x)= (\alpha-1) \sum_{j=1}^m p_j^{\alpha} -\alpha p_x^{\alpha-1}.
\end{equation}

To solve the constrained problem in equation \ref{eq:argmax}, we can apply the Lagrange multiplier:
\begin{equation}
f(p,\lambda, \mu) = p^\top z + H^\top_{\alpha}(p) - \lambda (\sum_j p_j - 1) - \sum_j \mu_j p_j.
\end{equation}
\begin{equation}
\frac{\partial f(p,\lambda, \mu)}{\partial p_j}=z_j - \frac{1}{\alpha-1}p_j^{\alpha-1} - \lambda - \mu_j = 0.
\end{equation}

Due to the complementary slackness condition of the KKT, if the solution $p_j>0$, then we have $\mu_j=0$, which yields:
\begin{equation}
z_j= \lambda + \frac{p_j^{\alpha-1}}{\alpha-1}.
\end{equation}
Similarly, if the probability of ground truth word $p_x > 0$, then $z_x= \lambda + \frac{p_x^{\alpha-1}}{\alpha-1}$. Substituting these into equation \ref{eq:entmaxloss}, we obtain:
\begin{equation}
\begin{aligned}
\mathcal{L}_{\alpha}(z,x)&=\sum_{j,p_j>0}p_jz_j - z_x + H^\top_{\alpha}(p)=\sum_{j, p_j>0}p_j(\lambda + \frac{1}{\alpha-1}p_j^{\alpha-1}) - \lambda - \frac{p_x^{\alpha-1}}{\alpha-1} + H^\top_{\alpha}(p)\\
&=\sum_j \frac{p_j^\alpha}{\alpha-1} - \frac{p_x^{\alpha-1}}{\alpha-1} + \frac{1}{\alpha(\alpha-1)}-\sum_j \frac{p_j^{\alpha}}{\alpha(\alpha-1)}\\
&=\frac{1}{\alpha(\alpha-1)}[(\alpha-1)\sum_j p_j^\alpha - \alpha p_x^{\alpha-1} + 1]\\
&=\frac{\mathcal{L}_{\alpha\text{-power}}(p,x) + 1}{\alpha(\alpha-1)}.
\end{aligned}
\end{equation}

As shown, the $\alpha$-entmax loss is a linear transformation of the $\alpha$-power score based loss, so they are fundamentally equivalent. This reveals the propriety of the $\alpha$-entmax loss. However, the equivalence does not hold when $p_x=0$. In this case, $z_x$ falls below the threshold of obtaining positive probability, causing the gradient from the probability vector $\frac{\partial p}{\partial z_x}$ to be 0. This makes it theoretically impossible to obtain a gradient from probability-based loss functions. Therefore, when applying other strictly proper scoring rules to the training of sparse transformations, adjustments are still necessary to ensure that the gradient can be transmitted to the golden logit $z_x$.

\section{Scoring Rules as Beam Search Objective}
We are grateful for the very insightful comments provided by Reviewer z8jq, which inspired us to investigate the application of scoring rules as objectives for beam search.

In the realm of conditional generation tasks such as machine translation and text summarization, beam search is a widely adopted decoding strategy aimed at finding the output sequence $y$ with the highest length-normalized log-probability. The formal objective of beam search can be expressed as:
\begin{equation}
\max_{y}\ \frac{\sum_{t=1}^{|y|}\log p_{\theta}(y_t|x,y_{<t})}{|y|^{\alpha}},
\end{equation}
where $\alpha$ denotes the length penalty hyperparameter. The above equation can also be understood as maximizing the sum of token-level logarithmic scores. Similarly, we can consider having beam search optimize other token-level scoring rules:
\begin{equation}
\max_{y}\ \frac{\sum_{t=1}^{|y|}S(p_{\theta}(\cdot|x,y_{<t}), y_t)}{|y|^{\alpha}}.
\end{equation}
Here, $S$ can be the Brier score $S(p,i)=2p_i-\sum_{j=1}^{m}p_j^2$, the spherical score $S(p,i)=\frac{p_i}{|p|}$, or other strictly proper scoring rules. A critical aspect is the sign (positive or negative) of the scoring rule. Given their definitions, the logarithmic score is inherently negative, the spherical score is positive, and the sign of the Brier score is uncertain. For a negative score like the logarithmic score, models tend to favor shorter sentences, whereas the length penalty $\alpha$ can counterbalance this by encouraging longer output. Conversely, for a positive score like the spherical score, models are inclined to generate longer sentences, and here, the length penalty $\alpha$ serves to encourage shorter sentences. To unify them, we subtract 1 from both the Brier score and the spherical score to ensure they are non-positive:
\begin{equation}
S_{Brier}'=2p_i-\sum_{j=1}^{m}p_j^2-1 \leq 2p_i - p_i - 1 \leq 0,\quad S_{Spherical}'=\frac{p_i}{|p|} - 1 \leq 1-1 =0.
\end{equation}

We conduct experiments on the WMT14 En-De dataset to evaluate the impact of different scoring rules on the quality of generated text when used as the objective for beam search. The results are presented in Table \ref{table:Results_decode}. The results indicate that, among the three scoring rules examined, the logarithmic score yields the best performance, with the Brier score outperforming the spherical score. However, there are exceptions. For instance, the model fine-tuned with the spherical score demonstrated a preference for beam search optimization using the spherical score over the Brier score.

\begin{table}[h]
\caption{BLEU scores on WMT14 En-De when applying different scoring rules as beam search objective.}
\label{table:Results_decode}
\vskip 0.1in
\begin{center}
\begin{small}
\begin{tabular}{lcccc}
\toprule
{\bf Model}  &  \bf Logarithmic & \bf Brier & \bf Spherical \\
\midrule
Transformer  & 27.61 & 27.56 & 27.23 \\
Transformer + Brier &   28.01 &   27.95 &  27.53 \\
Transformer + Spherical &   28.07 &  27.40 & 27.78 \\
\bottomrule 
\end{tabular}
\end{small}
\end{center}
\end{table}

Our investigation into the use of different scoring rules as objectives for beam search is far from exhaustive. It is plausible that other strictly proper scoring rules could surpass the performance of logarithmic score. We leave this for future exploration.
\end{document}